\documentclass[runningheads]{llncs}

\usepackage{eccv}

\usepackage{eccvabbrv}

\usepackage{graphicx}
\usepackage{booktabs}

\usepackage[accsupp]{axessibility}  

\usepackage{hyperref}

\usepackage{orcidlink}

\usepackage{multirow}
\usepackage{tabu}
\usepackage{pifont}
\usepackage{makecell}
\usepackage{amsmath}
\usepackage{array}
\usepackage[table]{xcolor}
\usepackage{subcaption}
\usepackage{wrapfig}
\usepackage{placeins}
\usepackage[dvipsnames]{xcolor}
\usepackage{float}

\newcommand{\xmark}{\ding{55}}
\newcommand{\cmark}{\ding{51}}
\newcommand{\mname}{}

\newcommand{\pubbacksz}{\fontsize{7.0}{7.0}\selectfont}
\newcolumntype{P}{>{\pubbacksz}c}

\newcommand{\pub}[1]{$({\tiny\textcolor{Purple}{\text{#1}}})$}

\colorlet{bapbg}{blue!12}

\usepackage{algorithm}
\usepackage{algorithmicx}
\usepackage{algpseudocode}
\usepackage{bm}

\title{Rethinking Prototype-based Similarity \\ Learning for Few-Shot Object Detection}
\titlerunning{ReSet}

\author{
KunHo Heo$^{*}$, 
Seungjae Kim$^{*}$, 
Wongyu Lee$^{*}$, 
SuYeon Kim, \\
and MyeongAh Cho$^{\dagger}$
}

\authorrunning{K.~Heo \textit{et al}.}

\institute{Kyung Hee University, Republic of Korea\\
\email{\{hkh7710, tmdwo8814, fovert, spoiuy3, maycho\}@khu.ac.kr}}

\begin{document}
\maketitle

\begingroup
\renewcommand\thefootnote{}
\footnotetext{* Equal contribution}
\footnotetext{\dag\ Corresponding author}
\endgroup

\vspace{-0.2cm}
\begin{abstract}
  Few-shot object detection aims to detect novel object categories from only a few labeled examples, avoiding costly large-scale annotation. Recent prototype-based similarity learning approaches enable training-free adaptation by matching query features with class prototypes. However, they suffer from two fundamental limitations: (i) \textbf{class confusion} arising from inter-class similarity margin collapse, and (ii) \textbf{insufficient visual cues} for precise localization, as similarity scores capture only class-level semantic affinity while providing limited spatial information. To address these issues, we introduce two complementary components. \textbf{Text-Anchored Semantic Mask (TSMa)} leverages class-level text features as semantic anchors to identify semantically aligned channels through channel-wise interaction between visual and text features. By suppressing style-induced spurious responses and  emphasizing class-intrinsic signals, TSMa enlarges inter-class similarity margins and mitigates class confusion. We further propose \textbf{Stage-Aligned Hierarchical Autoregressive Regression (SHARe)}, which reformulates localization as a hierarchical autoregressive process that progressively refines bounding boxes across multiple stages. SHARe leverages the layer-wise characteristics of ViT representations by aligning feature abstraction levels with regression stages: deeper layers guide early coarse localization, while shallower layers rich in edge and texture cues refine spatial details in later stages. Experiments on COCO demonstrate a new state of the art, outperforming the previous best by \textbf{+10.1 nAP}, with extensive analysis validating each component. The code is available at \url{https://github.com/VisualScienceLab-KHU/ReSet}.
\end{abstract}    
\section{Introduction}

Object detection is a fundamental problem in computer vision \cite{ren2015faster, he2017mask, carion2020end, caron2021emerging}, aiming to jointly localize objects and classify their semantic categories within an image. Despite remarkable progress in recent years, state-of-the-art detectors heavily rely on large-scale datasets with exhaustive bounding-box annotations. In practical applications, however, novel object categories continuously arise, making it prohibitively expensive and often infeasible to collect sufficient labeled samples for every emerging class. As a practical solution to this limitation, Few-shot Object Detection (FSOD) \cite{xin2024few, antonelli2022few} has emerged as a promising paradigm that enables the detection of novel classes from only a few annotated examples.

\begin{figure}[t]
    \centering
    \includegraphics[width=1\linewidth]{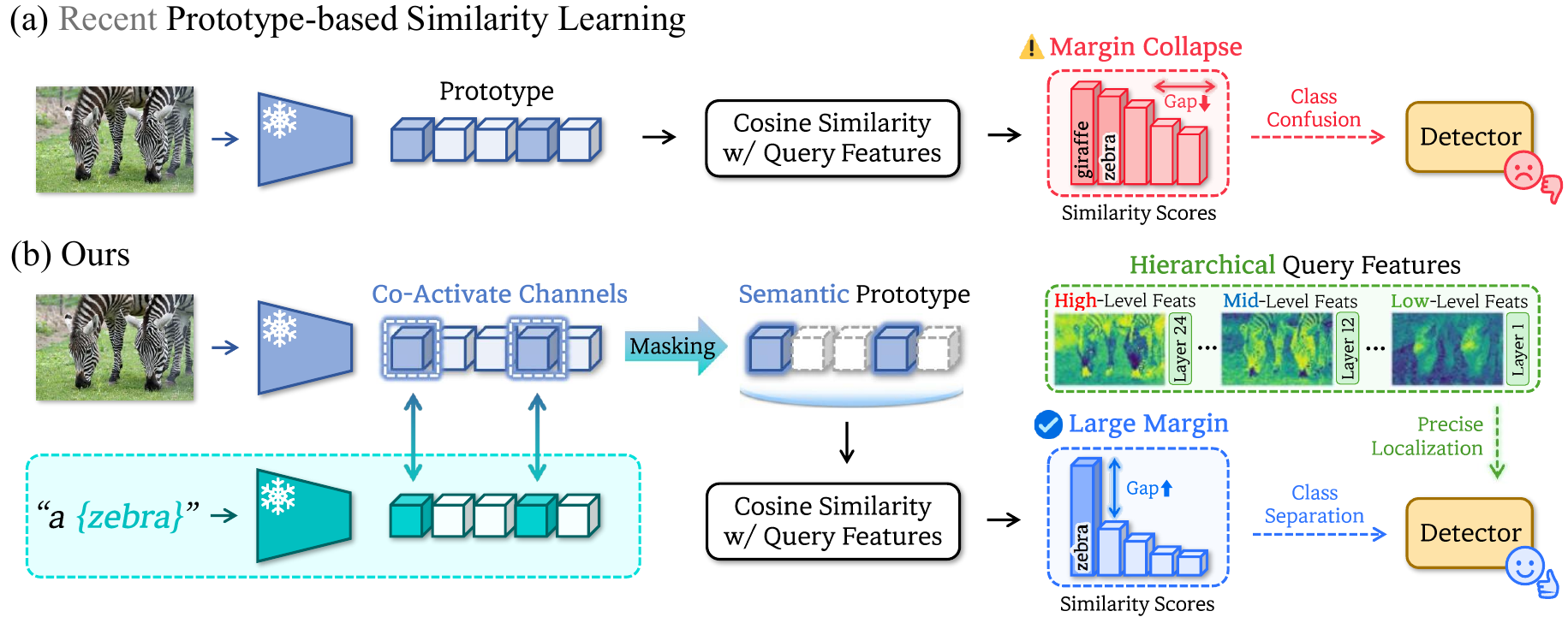}
    \vspace{-0.5cm}
    \caption{(a) Recent prototype-based similarity learning methods construct class prototypes solely from visual features, and similarity scores computed with query features can cause inter-class margin collapse, leading to class confusion. (b) Our method constructs \textit{Semantic Prototypes} by leveraging text features as anchors to retain only co-activated channels, enlarging inter-class similarity margins and thereby improving class separability. Additionally, multi-level hierarchical ViT features are injected to compensate for insufficient visual cues, enabling precise localization.}
    \vspace{-0.3cm}
    \label{fig:figure1}
\end{figure}

Early FSOD approaches \cite{fan2021generalized, qiao2021defrcn, sun2021fsce, wang2020frustratingly} first train a detector on \emph{base classes} with abundant annotations and then fine-tune it on \emph{novel classes} using a small \emph{support set}. However, such fine-tuning-based approaches often suffer from training instability and overfitting due to the scarcity of support samples, limiting generalization to novel categories. To mitigate this issue, recent works~\cite{zhang2023detect, fu2024cross, zhou2025pixel} adopt a training-free, prototype-based similarity learning that handles novel classes at inference time without additional fine-tuning. Specifically, class prototypes are constructed by extracting and aggregating features from support images. Given a query image, dense features are obtained using a ViT backbone \cite{oquab2023dinov2}, and class-wise similarity scores are computed (e.g., dot products) between query features and class prototypes to guide both classification and localization. DE-ViT~\cite{zhang2023detect} and PiDiViT~\cite{zhou2025pixel} are representative examples of this similarity-based learning approach. While they achieve strong performance without fine-tuning, they share two fundamental limitations.


The first limitation is \textbf{class confusion caused by inter-class similarity margin collapse.} Methods under the prototype-based similarity learning~\cite{zhang2023detect, zhou2025pixel} classify objects by leveraging similarity scores between query features and class prototypes, where the relative separation among these scores plays a crucial role. When multiple classes produce comparable scores, leading to inter-class margin collapse, as shown in Fig.~\ref{fig:figure1} (a), the model struggles to establish clear decision boundaries, resulting in systematic misclassification that propagates through subsequent detection stages. The second limitation is \textbf{insufficient visual cues for precise localization.} 
In prototype-based similarity learning, localization is also primarily conditioned on class-wise similarity scores. Since these scores are computed via cosine similarity and thus mainly encode class-level semantic affinity (mostly based on vector direction), they provide limited spatial information such as object boundaries, shapes, and scale, making precise localization inherently challenging.

To mitigate \textbf{(i) class confusion}, we introduce \textbf{Text-Anchored Semantic Mask (TSMa)}. Visual features extracted from support images inevitably capture not only class-discriminative cues but also style-related components shared across different classes. Such style-driven components introduce spurious similarity responses across semantically unrelated classes, leading to inter-class similarity margin collapse. In contrast, text features provide a compact representation of class semantics, are largely free from such style components and thus carry a higher proportion of class-intrinsic information. Motivated by this property, as shown in Fig.~\ref{fig:figure1} (b), we leverage text-derived class features as semantic anchors and identify semantically aligned channels via channel-wise interaction between visual and text features. The channels that are not jointly activated are then masked out, ensuring that similarity computation is dominated by semantically consistent channels. By amplifying contributions from class-intrinsic channels while attenuating style-induced responses, TSMa enlarges inter-class similarity margin and mitigates class confusion, as empirically validated in Sec.~\ref{sec:discussions}.

To address \textbf{(ii) lack of visual cues for localization}, we propose \textbf{Stage-Aligned Hierarchical Autoregressive Regression (SHARe)}. In existing prototype-based similarity learning, localization is performed in a single step conditioned on similarity scores, which provide limited structural and boundary information. Instead, SHARe reformulates localization as a hierarchical autoregressive process that refines bounding boxes across multiple stages. As shown in Fig.~\ref{fig:figure1} (b), we further leverage the layer-wise property of the ViT backbone representations \cite{dorszewski2025colors, kim2025interpreting} by injecting features of varying levels into each regression stage: deeper layers, which encode high-level semantics and global context, guide early coarse localization, while shallower layers, rich in low-level cues such as edges and textures, are progressively introduced to refine spatial details in later stages. This stage-aligned feature injection compensates for the representational limitations of similarity scores and enables precise object localization.

We evaluate our approach on the COCO benchmark~\cite{lin2014microsoft} under one-shot and 10/30-shot settings reporting novel-class performance (nAP/nAP50/nAP75). Our method establishes a new state of the art, outperforming the previous best by +10.1 nAP, +7.1 nAP50, and +9.8 nAP75 in the 30-shot setting. Furthermore, extensive analysis and ablation studies provide systematic evidence for the effectiveness of each component and design rationale, offering a strong and reproducible baseline for prototype-based similarity learning.
\begin{itemize}
    \item We identify class confusion induced by \emph{inter-class similarity margin collapse} as a key bottleneck in prototype-based similarity learning, and propose \textbf{Text-Anchored Semantic Mask (TSMa)} that uses text features as semantic anchors to suppress style-induced spurious similarity.
    \item We propose \textbf{Stage-Aligned Hierarchical Autoregressive Regression (SHARe)}, which performs progressive box refinement via stage-aligned injection of hierarchical ViT features, enabling precise localization.
    \item We achieve a new state of the art on COCO benchmark with substantial gain over the previous SOTA (e.g. +10.1 nAP), and validate our contributions through rigorous analysis, establishing a strong baseline for FSOD.
\end{itemize}
\section{Related Work}

\textbf{Few-Shot Object Detection (FSOD).} FSOD aims to detect novel categories from only a handful of labeled examples, leveraging knowledge transferred from base categories with abundant annotations \cite{antonelli2022few, kohler2023few, xin2024few}. Most benchmarks follow a two-stage protocol that trains a detector on base classes and then adapts it to disjoint novel classes under a few-shot setting \cite{wang2020frustratingly}, evaluated in a generalized setting that includes both base and novel classes. Existing methods can be broadly grouped into finetuning-based and meta-learning-based approaches. Finetuning-based methods \cite{fan2021generalized, sun2021fsce, wang2020frustratingly, guirguis2023niff, xiao2022few, qiao2021defrcn} adapt a pretrained detector to novel data in a simple and practical manner, but they are sensitive to training conditions and prone to overfitting toward base classes, leading to degraded novel class performance. Meta-learning methods train episodically across tasks to enable rapid adaptation. Early works rely on prototype-based matching \cite{kang2019few, yan2019meta, fan2020few}, while later designs introduce richer support–query interactions such as dense feature interaction or transformer conditioning \cite{han2021query, han2022few, han2022meta, zhang2022meta, bulat2023fs} to improve adaptability. These advances come at the cost of higher computation and potential amplification of support-set bias in low-shot regimes. Recently, the field has increasingly leveraged the rich representations of large pretrained vision transformers (ViTs) \cite{oquab2023dinov2}, often keeping the backbone frozen and shifting adaptation to inference-time mechanisms based on prototype-based similarity scores rather than heavy fine-tuning. Representative methods such as DE-ViT \cite{zhang2023detect} and PiDiViT \cite{zhou2025pixel} demonstrate that freezing the backbone and relying on prototype-based similarity scores can simultaneously alleviate the base-biased overfitting of finetuning-based methods and the computational overhead of meta-learning approaches. This prototype-based similarity learning has emerged as a novel direction in modern FSOD.

\noindent\textbf{Prototype-based Similarity Learning.} Early few-shot detectors construct class prototypes from support examples and use them to modulate or classify query features \cite{yan2019meta, kang2019few}. Subsequent meta-learning methods improve support–query alignment through dense feature interactions \cite{han2021query, han2022few, han2022meta, zhang2022meta, bulat2023fs}, yet at the cost of heavy computation and often benefiting from additional finetuning to achieve competitive novel-class accuracy. Departing from both lines, a recent paradigm keeps a pretrained ViT backbone \cite{caron2021emerging, oquab2023dinov2} and instead leverages prototype-based similarity scores as the core detection representation. DE-ViT~\cite{zhang2023detect} formalizes this direction by projecting query ViT features onto a prototype-spanned subspace via dot-product similarity and localizing objects via iterative region propagation rather than direct coordinate regression. PiDiViT~\cite{zhou2025pixel} inherits this pipeline and observes that frozen ViT features exhibit blurred boundary responses and excessively smooth center-to-boundary transitions, degrading object–background distinction. It mitigates these issues through pixel difference convolutions for boundary sharpening and multiscale feature fusion for scale-aware detection. Beyond in-domain benchmarks, this paradigm has also been extended to cross-domain few-shot detection with domain-aware adaptations such as instance features and domain prompting \cite{fu2024cross}, demonstrating its broad applicability across diverse target domains. Nevertheless, prototype-based similarity learning methods still exhibit two structural bottlenecks. First, visual prototypes inevitably encode style-related and instance-specific components shared across classes. These spurious components introduce similarity responses across semantically unrelated categories, causing inter-class similarity margin collapse and systematic class confusion. Second, similarity scores primarily encode class-level semantic affinity and provide limited geometric cues (e.g., boundaries, shape, and scale), making precise localization challenging. In this paper, we aim to address both bottlenecks of the prototype-based similarity learning by improving the reliability of similarity computation itself to mitigate inter-class confusion and by progressively leveraging hierarchical representations of the frozen ViT backbone \cite{dorszewski2025colors, kim2025interpreting} to supplement the missing geometric cues.

\section{Method}

\subsection{Overview}

\begin{figure}[t]
    \centering
    \includegraphics[width=1\linewidth]{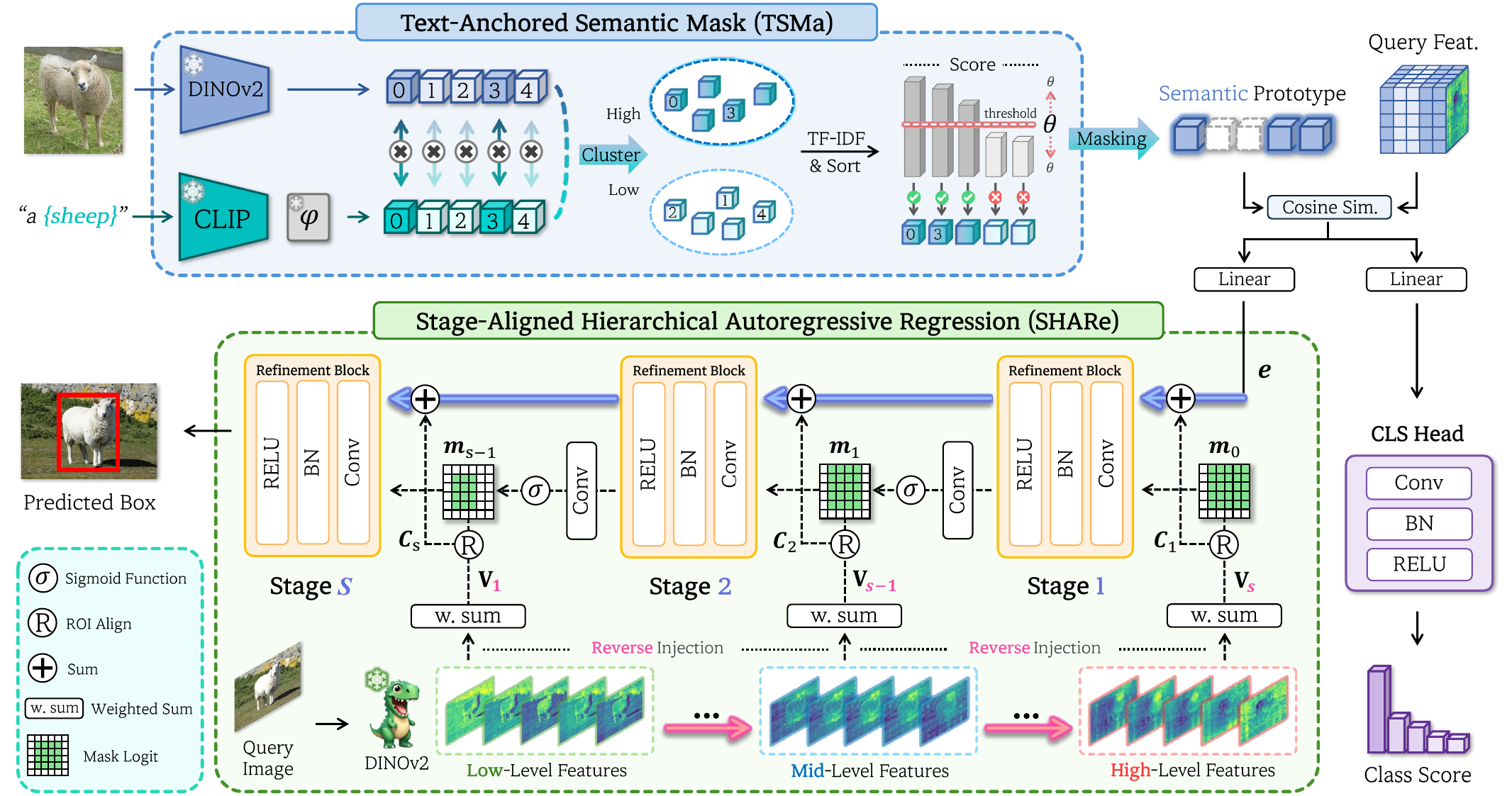}
    \caption{Overview of the proposed framework. It consists of two main components: \textbf{Text-Anchored Semantic Mask (TSMa)}, which constructs semantic prototypes by leveraging text embeddings as anchors, and \textbf{Stage-Aligned Hierarchical Autoregressive Regression (SHARe)}, which injects multi-level ViT features in reverse order into each autoregressive regression stage for precise localization.}
    \label{fig:figure2}
\end{figure}

As illustrated in Fig.~\ref{fig:figure2}, the proposed framework integrates \textbf{Text-Anchored Semantic Mask (TSMa)} and \textbf{Stage-Aligned Hierarchical Autoregressive Regression (SHARe)} into a unified pipeline. During training, a class prototype is constructed by averaging the visual features of object instances from each \textit{base} class, and a class-specific semantic mask generated via \textbf{TSMa} from the same \textit{base} class samples is applied to it to produce the \emph{Semantic Prototype}. Subsequently, the semantic prototypes and query features are normalized and their similarity is computed through cosine similarity, then passed through a linear projection to produce a \textit{similarity embedding} shared across two parallel branches: classification and regression. In the classification branch, the similarity embedding is fed into a classification head to produce a per-class score. In the regression branch, a mask logit corresponding to an initial RPN~\cite{ren2015faster} proposal is fed into the first stage together with the similarity embedding, and progressively refined across subsequent stages in an autoregressive manner via \textbf{SHARe}. The resulting refined mask is converted to a bounding box and combined with the per-class score to produce the final detection output. Training losses comprise a focal loss for per-class scoring and $\ell_1$ and GIoU losses for bounding box regression, supplemented by auxiliary BCE and Dice losses for mask refinement supervision. At inference, prototypes and semantic masks are constructed solely from the $k$-shot support examples of \textit{novel} classes without any fine-tuning, and the remaining pipeline proceeds identically.




\subsection{Text-Anchored Semantic Mask (TSMa)}
The core premise of \textbf{TSMa} is that channels co-activated by both visual and text features more directly reflect class-specific semantics. Since the visual backbone DINOv2~\cite{oquab2023dinov2} and the text backbone CLIP~\cite{radford2021learning} operate in distinct feature spaces, we first learn a lightweight mapping function that projects CLIP text features into the DINOv2 feature space.

\noindent\textbf{(1) Text-to-Vision Feature Mapping.}
We adopt the nonlinear warping proposed in Talk2DINO~\cite{barsellotti2025talking}, which maps CLIP text features into the DINOv2 patch feature space. Both backbones are frozen, and only the mapping function $\psi(\cdot)$ is trained. Specifically, $\psi(\cdot)$ consists of two affine transformations with a $\tanh$ nonlinearity:
\begin{equation}
    \psi(\mathbf{t}) = \mathbf{W}_b^\top \left(\tanh(\mathbf{W}_a^\top \mathbf{t} + \mathbf{b}_a)\right) + \mathbf{b}_b.
\end{equation}
We use COCO~\cite{lin2014microsoft} for training, extracting a text feature $\mathbf{t}$ for each class via a text prompt of the form \texttt{"\{class\}"}. The mapping function is optimized via contrastive learning~\cite{oord2018representation}, using the similarity between $\psi(\mathbf{t})$ and per-head region summary features obtained from DINOv2's self-attention as supervision.

\noindent\textbf{(2) Channel Clustering and Scoring.}
For class $c$, let $\tilde{\mathbf{t}}_c = \psi(\mathbf{t}_c)\in \mathbb{R}^D$ denote the mapped text feature, where $\psi$ is a pretrained mapping function and $D$ is the feature dimension. Given the visual feature $\mathbf{v}_{c,n}\in \mathbb{R}^D$ of its $n$-th object instance, we measure the per-channel semantic alignment via element-wise product:
\begin{equation}
    \mathbf{a}_{c,n} = \frac{\mathbf{v}_{c,n}}{|\mathbf{v}_{c,n}|_2} \odot \frac{\tilde{\mathbf{t}}_c}{|\tilde{\mathbf{t}}_c|_2} \in \mathbb{R}^D.
\end{equation}
A large value of $\mathbf{a}_{c,n}[i]$ indicates that the $i$-th channel responds strongly to both the visual and text feature. We apply $k$-means clustering ($k=2$) on the channels of $\mathbf{a}_{c,n}$, designating the higher-mean cluster as semantically-aligned active channels. This assignment is encoded as an indicator $g_{c,n}(i) \in \{0, 1\}$, where $g_{c,n}(i) = 1$ if channel $i$ belongs to the higher-mean cluster and $0$ otherwise. The activation frequency of channel $i$ for class $c$ is then aggregated over all $N_c$ instances as:
\begin{equation}
    \mathrm{count}_c(i) = \sum_{n=1}^{N_c} g_{c,n}(i).
\end{equation}

Subsequently, the importance of each channel is quantified as a score by adapting TF-IDF~\cite{manning2008introduction}:
\begin{gather}
    \mathrm{TF}_c(i) = \frac{\mathrm{count}_c(i)}{N_c}, \\
    \mathrm{DF}(i) = \frac{1}{|\mathcal{C}|} \sum_{c \in \mathcal{C}} \mathrm{TF}_c(i), \ \mathrm{IDF}(i) = \log\!\left(\frac{1}{\mathrm{DF}(i) + \varepsilon}\right), \\
    \mathrm{Score}_c(i) = \mathrm{TF}_c(i) \cdot \mathrm{IDF}(i).
\end{gather}
where $\mathcal{C}$ denotes the set of all classes. This formulation assigns high scores to channels that are frequently activated for class $c$ (high TF) yet rarely activated across other classes (high IDF), ensuring that retained channels are not only class-relevant but also class-discriminative, thereby yielding a semantically discriminative channel ranking. This scoring is performed independently for base and novel classes, where novel classes rely solely on $k$-shot support examples.

\noindent\textbf{(3) Learnable Thresholding and Masking.}
Let $r_{c,i}$ denote the rank of channel $i$ for class $c$ sorted by $\mathrm{Score}_c(i)$ in descending order. To avoid the discontinuity of hard thresholding, we employ a learnable rank-based mask:
\begin{equation}
    M_{c,i}(\theta_c) = \sigma\!\left(\frac{\theta_c - r_{c,i}}{\tau}\right),
\end{equation}
where $\sigma(\cdot)$ is the sigmoid function and $\tau$ controls the softness of the mask boundary. Channels with $r_{c,i} < \theta_c$ receive $M \approx 1$, those with $r_{c,i} > \theta_c$ receive $M \approx 0$, and boundary channels are assigned continuous weights in $(0, 1)$, forming a soft selection. The threshold $\theta_c$ is a learnable scalar initialized to the lowest rank, ensuring that nearly all channels are retained initially to prevent model collapse, with unnecessary channels progressively suppressed during training. At inference time, the threshold $\theta_c$ for novel classes is set to the mean of the thresholds learned over all base classes, requiring no additional tuning.

By applying the resulting semantic mask to the class prototype $\mathbf{p}_c \in \mathbb{R}^D$, we obtain a \textit{Semantic Prototype} $\tilde{\mathbf{p}}_c$, defined as:
\begin{equation}
    \tilde{\mathbf{p}}_c = \frac{\mathbf{M}_c \odot \mathbf{p}_c}{|\mathbf{M}_c \odot \mathbf{p}_c|_2},
\end{equation}
which is then used to compute a similarity with the query feature via cosine similarity, yielding a \textit{similarity embedding} $\mathbf{e}$ used for both classification and localization. By ensuring that the similarity computation is governed by semantically-aligned channels, TSMa directly mitigates the inter-class similarity margin collapse that acts as a bottleneck in prototype-based similarity learning.


\subsection{Stage-Aligned Hierarchical Autoregressive Regression (SHARe)}
Although TSMa removes style components to obtain \textit{semantic prototypes}, the resulting \textit{similarity embedding} $\mathbf{e}$ still lacks the visual cue required for precise localization. Since cosine similarity is primarily sensitive to the angle between feature vectors, the similarity embedding $\mathbf{e}$ mainly encodes their directional (angular) alignment rather than geometric cues. To address this limitation, we leverage the hierarchical depth property of frozen ViT representations. At each localization stage, we utilize visual representations from different ViT layers, which enables precise localization.

\noindent\textbf{(1) Multi-layer Feature Aggregation of ViT.}
Let $\mathbf{X}^\ell \in \mathbb{R}^{D \times H \times W}$ denote the feature map from the $\ell$-th layer of a frozen ViT backbone, where $D$ is the feature dimension, $H \times W$ is the spatial resolution of the ViT patch grid, and $L$ is the number of ViT layers. To provide stage-wise visual cues for all $S$ localization stages, we split the ViT layers into $S$ stage-aligned groups, where $\{\mathcal{G}_s\}_{s=1}^{S}$ denotes the set of layer indices belonging to each group (larger $s$ corresponds to later ViT layers), and each group contains approximately $L/S$ indices. For instance, $\mathcal{G}_1 = \{0,\ldots,L/S-1\}$ and $\mathcal{G}_S = \{(S-1)\cdot L/S,\ldots,L-1\}$.
We then apply learnable weighted aggregation within each group to obtain the visual cue $\mathbf{V}_s$:
\begin{equation}
    \mathbf{V}_s = \sum_{\ell \in \mathcal{G}_s} \alpha_{s,\ell}\,\mathrm{LN}(\mathbf{X}^\ell),
    \qquad
    \alpha_{s,\ell}=\frac{\exp(w_{s,\ell})}{\sum_{j\in\mathcal{G}_s}\exp(w_{s,j})},
    \label{eq:eq9}
\end{equation}
where $\mathrm{LN}(\cdot)$ denotes channel-wise LayerNorm that reduces layer-wise scale discrepancy and $w_{s,\ell}$ are learnable weights within each group. The resulting set of stage-wise visual cue $\{\mathbf{V}_s\}_{s=1}^{S}$ will be used in the next step to condition the stage-wise refinement in a stage-aligned manner.

\noindent\textbf{(2) Stage-Aligned Hierarchical Autoregressive Regression.}
Given an initial \textit{similarity embedding} $\mathbf{e}$, \textbf{SHARe} performs localization by iteratively updating the $\mathbf{e}$ over $S$ stages, progressively incorporating stage-wise visual cues at each step.
For stage $t \in \{1,\ldots,S\}$, the \emph{refinement block} updates the embedding $\mathbf{e}_{t-1}$ to $\mathbf{e}_t$ through a sequence of Conv$\rightarrow$BatchNorm$\rightarrow$ReLU operations. The updated $\mathbf{e}_t$ is then passed through a $1\times1$ convolution followed by a sigmoid activation to produce a mask logit $\mathbf{m}_t$ representing the object region. 
To effectively utilize the stage-wise visual cue $\{\mathbf{V}_s\}_{s=1}^{S}$, we reinterpret the hierarchical refinement process as proceeding in the \emph{reverse order} of ViT representation levels across the layer depth.
ViT representations transition from low-level patterns to high-level semantics as layer depth increases. Specifically, earlier layers tend to capture fine-grained geometric cues such as edges and boundaries, whereas later layers increasingly encode high-level information such as semantic context \cite{dorszewski2025colors, kim2025interpreting}.
Leveraging this property, we design stage-aligned autoregression to utilize high-level visual cue in early stages and progressively incorporate lower-level cue in later stages.
This enables SHARe to first coarsely localize the object using semantic context and then refine it by injecting geometric details for precise localization.
Accordingly, we align the stage progression with the hierarchy of visual cue in reverse order:
\begin{equation}
    (\text{stage }1 \rightarrow \text{stage }S)\ \Longleftrightarrow\ (\mathbf{V}_S \rightarrow \mathbf{V}_1).
\end{equation}
At stage $t$, the refinement is conditioned on the visual cue $\mathbf{V}_{\pi(t)}$ with $\pi(t)=S-t+1$, which is used to form a stage condition $\mathbf{C}_t$.
Specifically, we apply RoIAlign~\cite{he2017mask} on $\mathbf{V}_{\pi(t)}$ within the Region of Interest $\mathcal{R}$ to extract a region-specific visual cue, and then project it with a stage-specific linear layer to obtain $\mathbf{C}_t$:
\begin{equation}
    \mathbf{C}_t = \mathrm{Linear}_t\!\left(\mathrm{RoIAlign}\!\left(\mathbf{V}_{\pi(t)}, \mathcal{R}\right)\right).
\end{equation}
A single RoI $\mathcal{R}$ is shared across all stages, obtained by the initial RPN proposal.

Subsequently, the stage condition $\mathbf{C}_t$ is gated using the previous-stage mask logit $\mathbf{m}_{t-1}$ and injected into the latent embedding:
\begin{equation}
    \tilde{\mathbf{e}}_{t-1} = \mathbf{e}_{t-1} + \gamma_t \cdot \left(\mathbf{C}_t \odot \sigma(\mathbf{m}_{t-1})\right),
\end{equation}
where $\sigma(\cdot)$ denotes the sigmoid function, $\odot$ is element-wise product, and $\gamma_t$ is a learnable scalar that controls the amount of injected visual cues.
To carry forward mask predictions across stages, the previous mask logits are concatenated with the injected embedding to form the stage input:
\begin{equation}
    \hat{\mathbf{e}}_{t-1} = \mathrm{Concat}\!\left(\tilde{\mathbf{e}}_{t-1}, [\mathbf{m}_{t-1}, \ldots, \mathbf{m}_0]\right)
\end{equation}
where $\mathrm{Concat}(\cdot)$ concatenates tensors along the channel dimension. 
The refinement block then produces the updated embedding $\mathbf{e}_t$ from $\hat{\mathbf{e}}_{t-1}$, and predicts the corresponding mask logit $\mathbf{m}_t$. Finally, each stage's mask logit $\mathbf{m}_t$ is converted into a bounding box via the spatial integral layer~\cite{zhang2023detect}. More details are provided in the supplementary materials.

\section{Experiment}

\subsection{Dataset and Evaluation Metrics}
We evaluate on the COCO dataset~\cite{lin2014microsoft} following standard base/novel class splits~\cite{wang2020frustratingly, yang2022balanced}. Base classes are used for training, while novel classes are held out for evaluation. Novel class performance is measured using nAP, nAP50, and nAP75, and base class performance using bAP. For one-shot evaluation, we follow the conventional protocol~\cite{michaelis2018one} that divides the 80 COCO categories into four equal partitions (Split-1/2/3/4), using three partitions as base classes and the remaining one as novel classes. For Pascal VOC~\cite{everingham2010pascal}, we evaluate on Split-1/2/3 across 1/2/3/5/10-shot settings, reporting nAP50.

\subsection{Implementation Details}
We build on a two-stage detection framework~\cite{he2017mask}, adopting a pre-trained RPN~\cite{zhong2022regionclip} trained solely on base classes for proposal generation. The vision backbone is a frozen DINOv2~\cite{oquab2023dinov2} ViT-L, and text features for TSMa are extracted using a frozen CLIP~\cite{radford2021learning} ViT-L text encoder.
Background class prototypes are extracted from COCO-Stuff~\cite{caesar2018coco}. During evaluation, the model is tested on images containing objects of both base and novel classes, selecting the top-$T$ ($T$=5) categories based on the similarity between query features and class prototypes.

\begin{table}[h]\centering 
\caption{Results on COCO 2014 few-shot benchmark.}
\label{tab:table1}
\resizebox{\linewidth}{!}{
\scriptsize
\begin{tabular}{l l c c cccc cccc}
\toprule
{\multirow{2}{*}{Method}} & &
\multirow{2}{*}{Backbone} &
\multirow{2}{*}{Finetune} &
\multicolumn{4}{c}{10-shot} &
\multicolumn{4}{c}{30-shot} \\
\cmidrule(l{2pt}r{2pt}){5-8} \cmidrule(l{2pt}r{0pt}){9-12}
 & &  & &
\multicolumn{1}{c}{bAP} & nAP & nAP50 & nAP75 &
\multicolumn{1}{c}{bAP} & nAP & nAP50 & nAP75 \\
\midrule

\multicolumn{2}{l}{FSCE~\pub{CVPR'21}~\cite{sun2021fsce}} & RN101 & \textcolor{green!70!black}{\cmark} & {-} & 11.9 & - & 10.5 & {-} & 16.4 & - & 16.2 \\
\multicolumn{2}{l}{Retentive RCNN~\pub{CVPR'21}~\cite{fan2021generalized}} & RN101 & \textcolor{green!70!black}{\cmark} & 39.2 & 10.5 & 19.5 & 9.3 & 39.3 & 13.8 & 22.9 & 13.8 \\
\multicolumn{2}{l}{HeteroGraph~\pub{ICCV'21}~\cite{han2021query}} & RN101 & \textcolor{green!70!black}{\cmark} & {-} & 11.6 & 23.9 & 9.8 & {-} & 16.5 & 31.9 & 15.5 \\
\multicolumn{2}{l}{FsDetView~\pub{TPAMI'22}~\cite{xiao2022few}} & RN50 & \textcolor{green!70!black}{\cmark} & 6.4 & 7.6 & - & - & 9.3 & 12 & - & - \\
\multicolumn{2}{l}{Meta Faster RCNN~\pub{AAAI'22}~\cite{han2022meta}} & RN101 & \textcolor{green!70!black}{\cmark} & {-} & 12.7 & 25.7 & 10.8 & {-} & 16.6 & 31.8 & 15.8 \\
\multicolumn{2}{l}{LVC~\pub{CVPR'22}~\cite{kaul2022label}} & ViT-S/8 & \textcolor{green!70!black}{\cmark} & 28.7 & 19.0 & 34.1 & 19 & 34.8 & 26.8 & 45.8 & 27.5 \\
\multicolumn{2}{l}{CrossTransformer~\pub{CVPR'22}~\cite{han2022few}} & PVTv2-B2-Li & \textcolor{green!70!black}{\cmark} & {-} & 17.1 & 30.2 & 17 & {-} & 21.4 & 35.5 & 22.1 \\
\multicolumn{2}{l}{NIFF~\pub{CVPR'23}~\cite{guirguis2023niff}} & RN101 & \textcolor{green!70!black}{\cmark} & 39.0 & 18.8 & - & - & 39.0 & 20.9 & - & - \\
\multicolumn{2}{l}{DiGeo~\pub{CVPR'23}~\cite{ma2023digeo}} & RN101 & \textcolor{green!70!black}{\cmark} & 39.2 & 10.3 & 18.7 & 9.9 & 39.4 & 14.2 & 26.2 & 14.8 \\
\midrule
\multicolumn{2}{l}{DE-ViT~\pub{CoRL'24}~\cite{zhang2023detect}} & ViT-L/14 & \textcolor{red!90!black}{\xmark} & 29.4 & 34.0 & 52.9 & 37.0 & 29.5 & 34.0 & 53.0 & 37.2 \\
\multicolumn{2}{l}{CD-ViTO~\pub{ECCV'24}~\cite{fu2024cross}} & ViT-L/14 & \textcolor{red!90!black}{\xmark} & {-} & 35.3 & 54.9 & 37.2 & {-} & 35.9 & 54.5 & 38.0 \\
\multicolumn{2}{l}{PiDiViT~\pub{ICCV'25}~\cite{zhou2025pixel}} & ViT-L/14 & \textcolor{red!90!black}{\xmark} & 32.3 & 37.0 & 57.6 & 40.9 & 31.4 & 37.3 & 58.5 & 41.7 \\
\rowcolor{bapbg}\multicolumn{2}{l}{\mname Ours} & ViT-L/14 & \textcolor{red!90!black}{\xmark} &
\textbf{41.2} & \textbf{47.1} & \textbf{65.5} & \textbf{51.5} &
\textbf{41.3} & \textbf{47.4} & \textbf{65.6} & \textbf{51.5} \\
\bottomrule
\end{tabular}
}
\end{table}

\subsection{Comparisons with SOTA Methods}


\noindent
\begin{wraptable}{r}{0.63\textwidth}
\vspace{-3.3em}
\centering
\caption{\small Results on COCO 2017 one-shot benchmark.}
\label{tab:table2}
\resizebox{\linewidth}{!}{%
\scriptsize
\begin{tabular}{l c cccc|c}
\toprule
\multirow{2}{*}{Method} & \multirow{2}{*}{Backbone} & \multicolumn{5}{c}{nAP50} \\
\cmidrule(l{0pt}r{0pt}){3-7}
 &  & Split-1 & Split-2 & Split-3 & Split-4 & \multicolumn{1}{c}{Avg} \\
\midrule
CoAE~\cite{hsieh2019one} & RN50     & 23.4 & 23.6 & 20.5 & 20.4 & 22.0 \\
AIT~\cite{chen2021adaptive} & RN50     & 26.0 & 26.4 & 22.3 & 22.6 & 24.3 \\
SaFT~\cite{zhao2022semantic} & RN101    & 27.8 & 27.6 & 21.0 & 23.0 & 24.9 \\
BHRL~\cite{yang2022balanced} & RN50     & 26.1 & 29.0 & 22.7 & 24.5 & 25.6 \\
DE-ViT~\cite{zhang2023detect} & ViT-L/14 & 27.4 & 33.2 & 27.1 & 26.1 & 28.4 \\
PiDiViT~\cite{zhou2025pixel} & ViT-L/14 & {32.0} & {38.6} & {30.4} & {30.2} & {32.8} \\
\midrule
\rowcolor{bapbg}{\mname Ours} & ViT-L/14 & \textbf{46.6} & \textbf{48.0} & \textbf{42.5} & \textbf{37.9} & \textbf{43.7} \\
\bottomrule
\end{tabular}%
}
\vspace{-1.0em}
\end{wraptable}
\noindent\textbf{Quantitative Evaluation.} Our method achieves substantial improvements over the previous methods including DE-ViT~\cite{zhang2023detect} and PiDiViT~\cite{zhou2025pixel}, establishing a new state of the art across diverse metrics and shot settings. As reported in Tab.~\ref{tab:table1}, we record gains of nAP +10.1, nAP50 +7.9, nAP75 +10.6, and bAP +8.9 in the 10-shot setting, and nAP +10.1, nAP50 +7.1, nAP75 +9.8, and bAP +9.9 in the 30-shot setting.
Tab.~\ref{tab:table2} further shows consistent gains across all four COCO one-shot splits, with an average nAP50 improvement of +10.9 over the previous SOTA. On Pascal VOC~\cite{everingham2010pascal}, Tab.~\ref{tab:table3} reports an average nAP50 gain of +6.9\% across all splits and shot settings. Taken together, these results firmly establish the superiority of our approach across two datasets and diverse shot conditions.

\begin{table*}[t]
\centering
\caption{nAP50 results on Pascal VOC few-shot benchmark.}
\label{tab:table3}
\small
\setlength{\tabcolsep}{3.2pt}
\renewcommand{\arraystretch}{1.05}
\resizebox{\textwidth}{!}{%
\begin{tabular}{l c *{5}{c} *{5}{c} *{5}{c} |c}
\toprule
\multirow{2}{*}{Method} & \multirow{2}{*}{Backbone} &
\multicolumn{5}{c}{Novel Split 1} &
\multicolumn{5}{c}{Novel Split 2} &
\multicolumn{5}{c}{Novel Split 3} &
\multicolumn{1}{|c}{\multirow{2}{*}{Avg}} \\
\cmidrule(lr){3-7}\cmidrule(lr){8-12}\cmidrule(lr){13-17}
& & 1 & 2 & 3 & 5 & 10 & 1 & 2 & 3 & 5 & 10 & 1 & 2 & 3 & 5 & 10 & \\
\midrule
TFA~\cite{wang2020frustratingly} & RN101 & 39.8 & 36.1 & 44.7 & 55.7 & 56.0 & 23.5 & 26.9 & 34.1 & 35.1 & 39.1 & 30.8 & 34.8 & 42.8 & 49.5 & 49.8 & 39.9 \\
Multi-Relation Det~\cite{fan2020few} & RN50 & 37.8 & 43.6 & 51.6 & 56.5 & 58.6 & 22.5 & 30.6 & 40.7 & 43.1 & 47.6 & 31.0 & 37.9 & 43.7 & 51.3 & 49.8 & 43.1 \\
Retentive RCNN~\cite{fan2021generalized} & RN101 & 42.4 & 45.8 & 45.9 & 53.7 & 56.1 & 21.7 & 27.8 & 35.2 & 37.0 & 40.3 & 30.2 & 37.6 & 43.0 & 49.7 & 50.1 & 41.1 \\
Meta Faster R-CNN~\cite{han2022meta} & RN101 & 43.0 & 54.5 & 60.6 & 66.1 & 65.4 & 27.7 & 35.5 & 46.1 & 47.8 & 51.4 & 40.6 & 46.4 & 53.4 & 59.9 & 58.6 & 50.5 \\
LVC~\cite{kaul2022label} & ViT-S/8 & 54.5 & 53.2 & 58.8 & 63.2 & 65.7 & 32.8 & 29.2 & 50.7 & 49.8 & 50.6 & 48.4 & 52.7 & 55.0 & 59.6 & 59.6 & 52.3 \\
CrossTransformer~\cite{han2022few} & PVTv2 & 49.9 & 57.1 & 57.9 & 63.2 & 67.1 & 27.6 & 34.5 & 43.7 & 49.2 & 51.2 & 39.5 & 54.7 & 52.3 & 57.0 & 58.7 & 50.9 \\
HeteroGraph~\cite{han2021query} & RN101 & 42.4 & 51.9 & 55.7 & 62.6 & 63.4 & 25.9 & 37.8 & 46.6 & 48.9 & 51.1 & 35.2 & 42.9 & 47.8 & 54.8 & 53.5 & 48.0 \\
DiGeo~\cite{ma2023digeo} & RN101 & 37.9 & 39.4 & 48.5 & 58.6 & 61.5 & 26.6 & 28.9 & 41.9 & 42.1 & 49.1 & 30.4 & 40.1 & 46.9 & 52.7 & 54.7 & 44.0 \\
NIFF~~\cite{guirguis2023niff} & RN101 & \textbf{62.8} & \textbf{67.2} & 68.0 & 70.3 & 68.8 & 38.4 & 42.9 & 54.0 & 56.4 & 54.0 & 56.4 & 62.1 & 61.2 & 64.1 & 63.9 & 59.4 \\
DE-ViT~\cite{zhang2023detect} & ViT-L/14 & 55.4 & 56.1 & 68.1 & 70.9 & 71.9 & 43.0 & 39.3 & 58.1 & {61.6} & {63.1} & 58.2 & 64.0 & 61.3 & {64.2} & 67.3 & 60.2 \\
PiDiViT~\cite{zhou2025pixel} & ViT-L/14 & {57.3} & {56.9} & {68.1} & {73.7} & {73.1} & {43.5} & {44.7} & {61.2} & 61.2 & 62.4 & {58.4} & {64.2} & {61.4} & 64.1 & {67.6} & {61.2} \\
\midrule
\rowcolor{bapbg}\mname Ours & ViT-L/14 & 59.2 & 64.3 & \textbf{75.6} & \textbf{79.5} & \textbf{79.7} & \textbf{50.8} & \textbf{56.0} & \textbf{67.6} & \textbf{67.7} & \textbf{67.3} & \textbf{65.5} & \textbf{72.0} & \textbf{68.9} & \textbf{72.8} & \textbf{74.6} & \textbf{68.1} \\
\bottomrule
\end{tabular}%
}
\vspace{-1.0em}
\end{table*}

\begin{figure}[H]
    \centering
    \includegraphics[width=\textwidth]{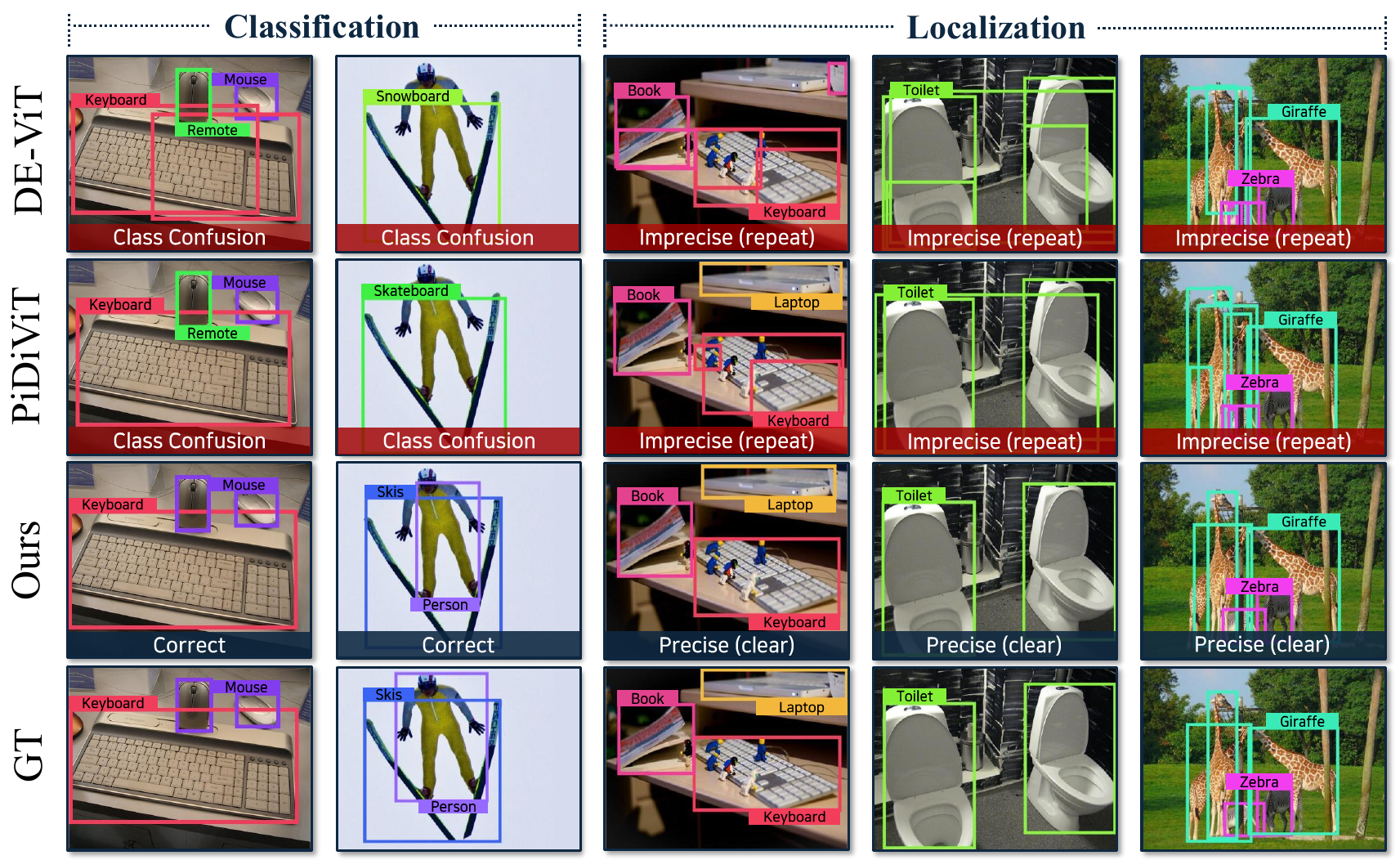}
    \caption{Qualitative comparison with existing methods (30-shot).}
    \label{fig:figure3}
    \vspace{-1.0em}
\end{figure}

\noindent\textbf{Qualitative Evaluation.} Fig.~\ref{fig:figure3} compares detection results of DE-ViT~\cite{zhang2023detect}, PiDiViT~\cite{zhou2025pixel}, and our method from classification and localization perspectives. \textbf{Classification}: in the first and second columns, DE-ViT and PiDiViT misclassify a \texttt{Mouse} as \texttt{Remote}, whereas our method correctly detects it. \textbf{Localization}: in the third through fifth columns, both baselines produce multiple redundant detections for a single object, while our method yields a single accurate prediction consistent with the ground truth. These results directly validate the contribution of each component: the classification improvements reflect \textbf{TSMa}'s ability to mitigate \textit{class confusion} from inter-class similarity margin collapse, while the localization improvements demonstrate \textbf{SHARe}'s effectiveness in compensating for \textit{insufficient visual cues} through hierarchical ViT feature injection. Additional qualitative samples are provided in the supplementary materials.

\begin{figure}[t]
    \centering
    \includegraphics[width=1\linewidth]{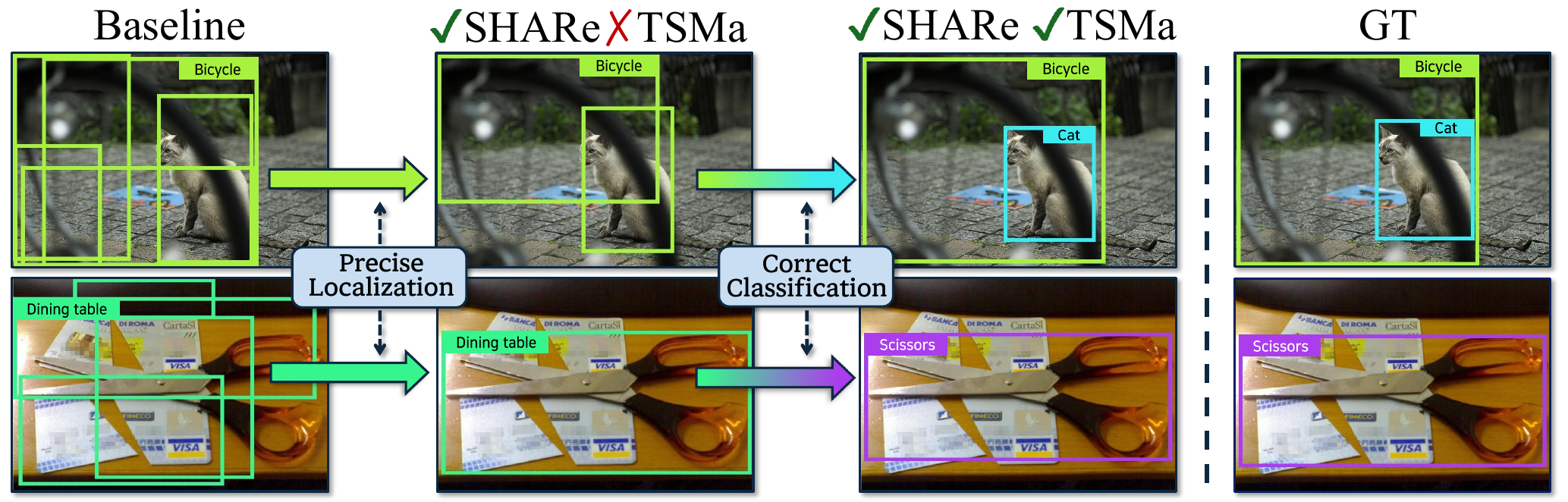}
    \caption{Visualization of ablation results on the proposed components.}
    \label{fig:figure4}
    \vspace{-0.5em}
\end{figure}

\subsection{Discussions}
\label{sec:discussions}
\noindent
\begin{wraptable}{r}{0.55\textwidth}
\vspace{-3.3em}
\centering
\caption{Ablation results on the proposed components.}
\label{tab:table4}

\setlength{\tabcolsep}{5pt}
\renewcommand{\arraystretch}{1.12}

\scriptsize
\begin{tabular}{c c | cccc}
\toprule
SHARe & TSMa & bAP & nAP & nAP50 & nAP75 \\
\midrule
-- & -- & 30.0 & 34.3 & 53.6 & 36.8 \\
\textcolor{black}{\cmark} & -- & 37.1 & 43.1 & 60.8 & 47.4 \\
-- & \textcolor{black}{\cmark} & 38.6 & 43.7 & 63.5 & 46.1 \\
\rowcolor{bapbg}\textcolor{black}{\cmark} & \textcolor{black}{\cmark} & \textbf{41.3} & \textbf{47.4} & \textbf{65.6} & \textbf{51.5} \\
\bottomrule
\end{tabular}
\vspace{-1.0em}
\end{wraptable}
\textbf{Ablation Studies on the Proposed Components.}
We present ablation results both quantitatively (Tab.~\ref{tab:table4}) and qualitatively (Fig.~\ref{fig:figure4}). SHARe alone yields consistent gains across all metrics, with the largest improvement on nAP75 (+10.6), which is most sensitive to localization precision at high IoU, suggesting that SHARe most prominently enhances \emph{localization}. This is further corroborated in Fig.~\ref{fig:figure4}, where redundant box detections observed in the baseline are resolved upon adding SHARe, demonstrating that injecting hierarchical ViT features from high-level semantics to low-level spatial cues effectively induces precise localization. TSMa alone likewise improves across all metrics, with the largest gain on nAP50 (+9.9), a metric more sensitive to classification under its lenient IoU criterion, suggesting that TSMa most prominently enhances \emph{inter-class discriminability}. Fig.~\ref{fig:figure4} further shows that misclassifications present in the SHARe-only model are corrected upon adding TSMa, demonstrating that constructing Semantic Prototypes by suppressing class-agnostic channels and emphasizing class-specific ones leads to improved inter-class separation.

\noindent
\begin{wraptable}{r}{0.45\textwidth}
\vspace{-3.3em}
\centering
\caption{Ablation results on TF-IDF scoring in TSMa.}
\label{tab:table5}

\small
\setlength{\tabcolsep}{4.2pt}
\renewcommand{\arraystretch}{1.12}

\resizebox{\linewidth}{!}{%
\scriptsize
\begin{tabular}{c | cccc}
\toprule
TF-IDF & bAP & nAP & nAP50 & nAP75 \\
\midrule
\textcolor{black}{\xmark} & 40.1 & 45.8 & 64.1 & 50.0 \\
\rowcolor{bapbg}\textcolor{black}{\cmark} & \textbf{41.3} & \textbf{47.4} & \textbf{65.6} & \textbf{51.5} \\
\bottomrule
\end{tabular}%
}
\vspace{-1.0em}
\end{wraptable}
\noindent\textbf{TF-IDF Scoring in TSMa.} 
In the TSMa pipeline, after partitioning channels into high/low-mean clusters via $k$-means clustering, we further refine per-channel scores through TF-IDF scoring, whose effect is reported in Tab.~\ref{tab:table5}. Among the per-channel magnitudes of the element-wise product between visual and text features, certain channels exhibit consistently large responses across all classes regardless of class identity, introducing confusion in inter-class discrimination. TF-IDF scoring mitigates this by penalizing such class-agnostic channels while emphasizing class-specific ones, and the performance drop when removing TF-IDF scoring corroborates this effect.

\begin{figure}[t]
    \centering
    \includegraphics[width=1\linewidth]{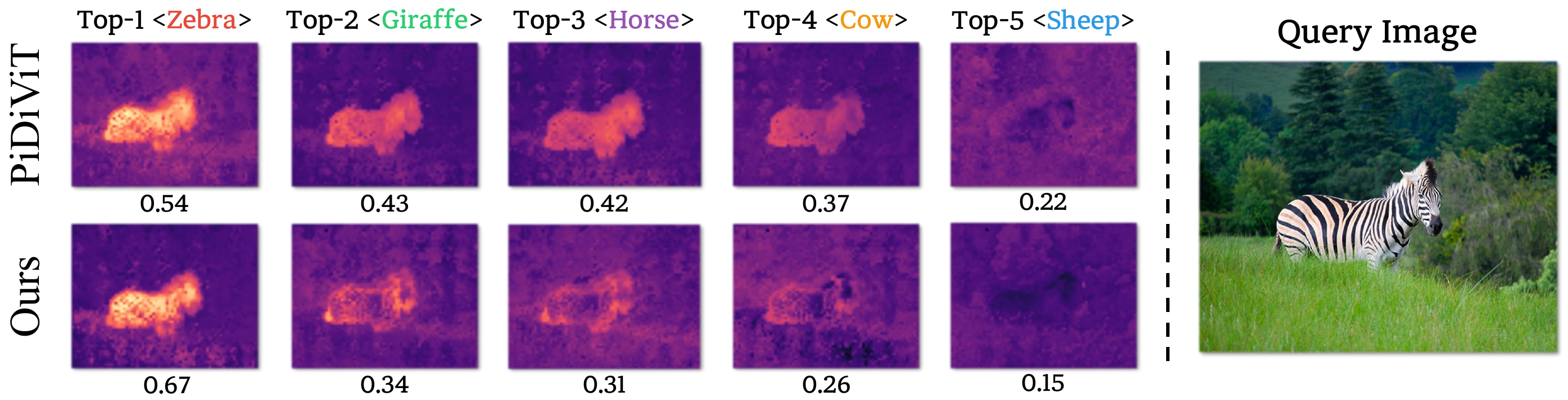}
    \caption{Query–prototype similarity heatmaps of Top-5 class prototypes.}
    \label{fig:figure5}
    \vspace{-1.5em}
\end{figure}

\noindent
\begin{wrapfigure}{r}{0.59\textwidth}
\vspace{-1.1em}
\centering
\begingroup
\captionsetup[table]{format=plain,width=0.92\linewidth,justification=centering,singlelinecheck=false}
\captionof{table}{Top-$k$ accuracy comparison.}
\label{tab:table6}
\endgroup
\resizebox{\linewidth}{!}{%
\scriptsize
\setlength{\tabcolsep}{6pt}
\renewcommand{\arraystretch}{1.10}
\begin{tabular}{l c c c c}
\toprule
\multirow{2}{*}{Method} & \multicolumn{3}{c}{Accuracy (\%)} & \multirow{2}{*}{nAP50} \\
\cmidrule(lr){2-4}
& Top-1 & Top-3 & Top-5 & \\
\midrule
PiDiViT~\cite{zhou2025pixel} & 63.2 & 72.1 & 87.2 & 58.5 \\
\rowcolor{bapbg}{\mname Ours} & \textbf{84.3} & \textbf{89.6} & \textbf{96.8} & \textbf{65.6} \\
\rowcolor{bapbg}
\hspace*{5pt}\textcolor{blue!80}{\scriptsize $\Delta$} &
\textcolor{blue!90}{(\scriptsize +21.1\%)} &
\textcolor{blue!90}{(\scriptsize +17.5\%)} &
\textcolor{blue!90}{(\scriptsize +9.6\%)} &
\textcolor{blue!90}{(\scriptsize +7.1\%)} \\
\bottomrule
\end{tabular}
}
\includegraphics[width=\linewidth]{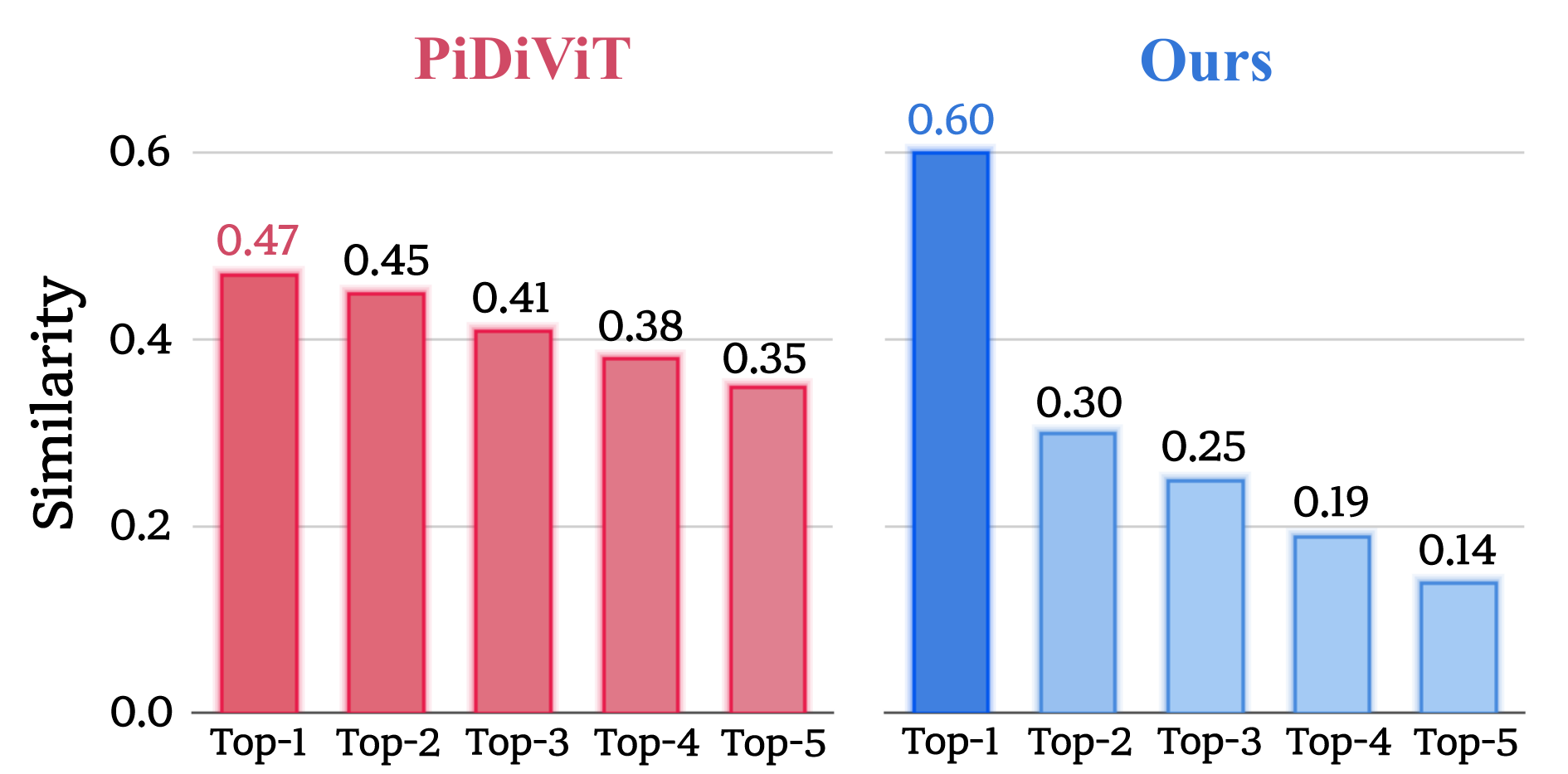}
\vspace{-1.9em}
\captionof{figure}{Comparison of average similarity scores.} 
\label{fig:figure6}
\vspace{0.4em}
\includegraphics[width=\linewidth]{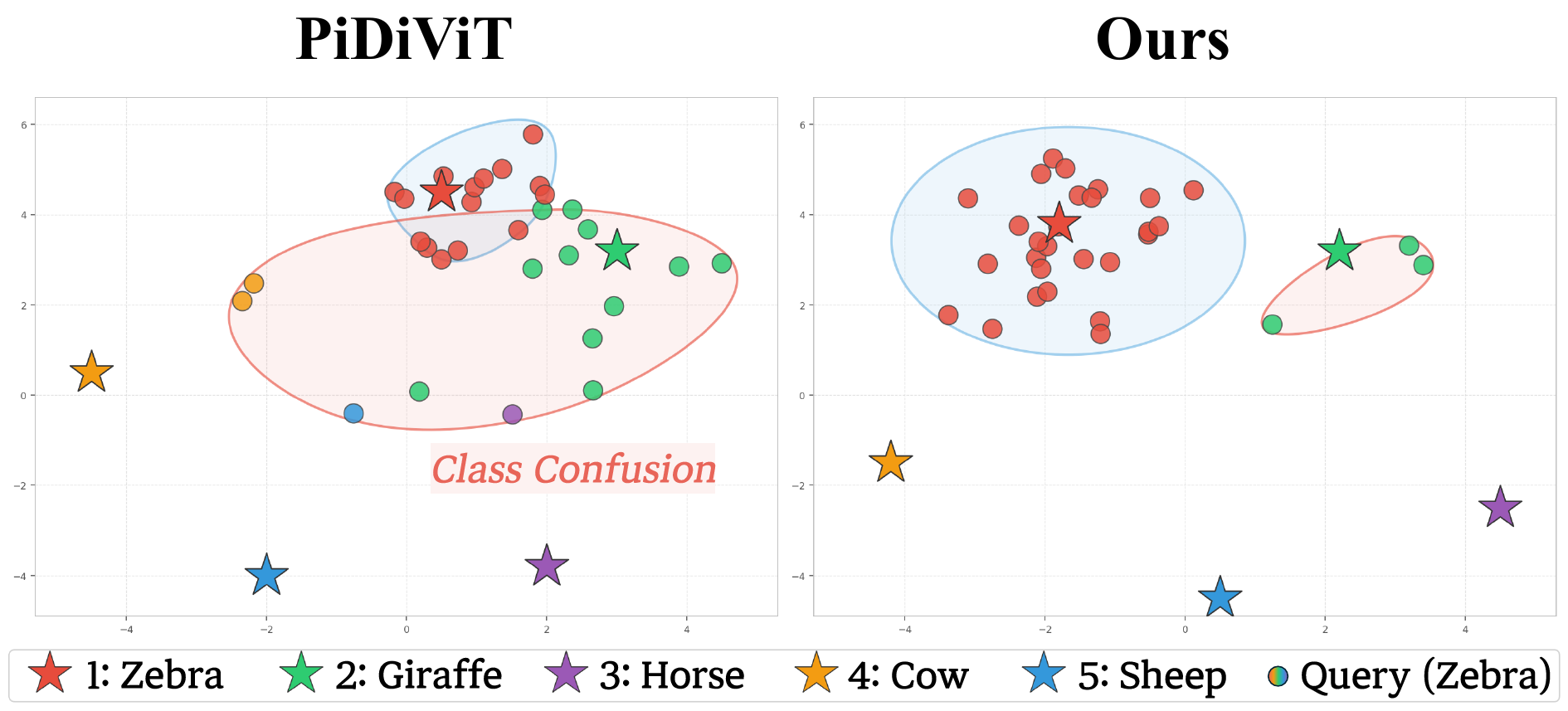}
\vspace{-1.9em}
\captionof{figure}{Comparison of t-SNE visualization.}
\label{fig:figure7}
\vspace{-1.0em}
\end{wrapfigure}
\textbf{Inter-Class Similarity Margin Enlargement via TSMa.}
To validate that TSMa resolves inter-class similarity margin collapse and improves class separability, we conduct analyses on the COCO~\cite{lin2014microsoft} test set. Tab.~\ref{tab:table6} reports Top-$k$ Accuracy, defined as the probability that the ground-truth class ranks among the top-$k$ most similar prototypes to the query feature, where our method outperforms PiDiViT by +21.1\%, +17.5\%, and +9.6\% at Top-1, Top-3, and Top-5, respectively. Fig.~\ref{fig:figure6} visualizes the average similarity scores of the top-5 class prototypes across all query features, showing substantially larger inter-class margins in our method. The enlarged Top-1/Top-2 gap indicates reduced confusion between the most confident class and its nearest competitor, while the wider margins across all top-5 classes corroborate improved inter-class discriminability. Fig.~\ref{fig:figure5} further illustrates this through similarity heatmaps: in PiDiViT, the gap between Top-1 and Top-2 is narrow, and Top-2 through Top-4 classes maintain high scores, whereas our method produces a pronounced gap beyond Top-1. Fig.~\ref{fig:figure7} presents t-SNE maps of 30 \texttt{Zebra} query features alongside the top-5 prototypes, where PiDiViT shows clear class confusion with many queries clustering near incorrect prototypes, while ours largely eliminates confusion. Collectively, these analyses confirm that TSMa successfully mitigates inter-class similarity margin collapse and improves class separability.

\begin{figure}[t]
    \centering
    \includegraphics[width=1\linewidth]{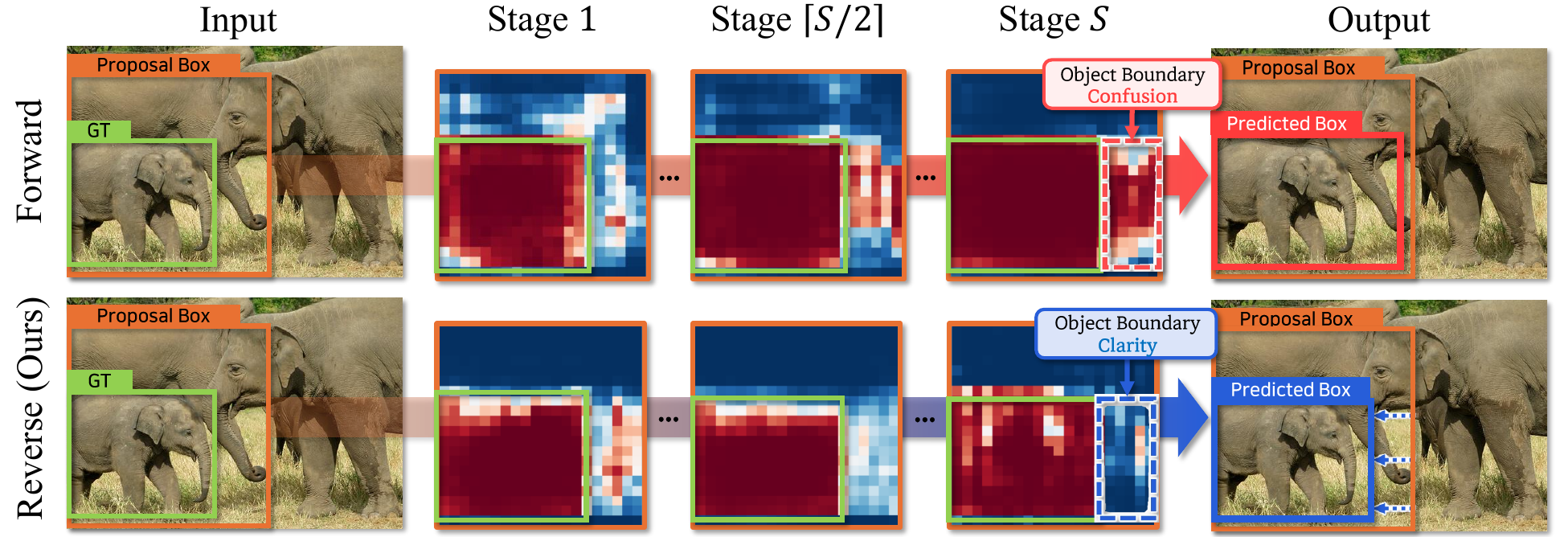}
    \caption{Visual comparison of ViT feature injection strategies in SHARe.}
    \label{fig:figure8}
    \vspace{-0.5em}
\end{figure}


\noindent
\begin{wraptable}{r}{0.45\textwidth}
\vspace{-1.1em}
\centering
\caption{Ablation results on injection strategies in SHARe.}
\label{tab:table7}

\small
\setlength{\tabcolsep}{4.4pt}
\renewcommand{\arraystretch}{1.12}

\resizebox{\linewidth}{!}{%
\begin{tabular}{l | cccc}
\toprule
Strategy & bAP & nAP & nAP50 & nAP75 \\
\midrule
None      & 38.6 & 43.7 & 63.5 & 46.1 \\
Forward   & 40.0 & 45.9 & 64.4 & 48.2 \\
\rowcolor{bapbg}\textbf{Reverse} & \textbf{41.3} & \textbf{47.4} & \textbf{65.6} & \textbf{51.5} \\
\bottomrule
\end{tabular}%
}
\vspace{-1.0em}
\end{wraptable}
\noindent\textbf{Ablations on SHARe Feature Injection.}
Stage-wise injection is central to SHARe, enabling the use of hierarchical ViT cues during autoregressive mask refinement. We report the effect of varying the injection order of ViT feature groups across stages in Tab.~\ref{tab:table7}. Compared to ``\texttt{None}'', ``\texttt{Forward}'' substantially improves all metrics, confirming that explicit stage-wise visual conditioning is crucial as similarity embeddings alone provide insufficient geometric cues. However, ``\texttt{Reverse}'' achieves the best performance across all metrics, with a particularly large gain on nAP75 (+3.3 over ``\texttt{Forward}''), demonstrating superior localization at high IoU. This validates our key insight that progressive localization is best aligned with the reverse of ViT's depth hierarchy: high-level semantic features establish coarse object extent in early stages, while lower-level spatial cues progressively sharpen boundaries in later stages. Fig.~\ref{fig:figure8} further corroborates this qualitatively; while ``\texttt{Forward}'' produces reasonable early-stage localization, it increasingly confuses overlapping instances as refinement progresses, whereas ``\texttt{Reverse}'' consistently resolves such confusion across stages.

\vspace{-0.5em}
\section{Conclusion}
We address two fundamental limitations of prototype-based similarity learning in FSOD: inter-class similarity margin collapse causing class confusion, and insufficient visual cues for precise localization. To mitigate the former, \textbf{TSMa} leverages text features as semantic anchors to identify semantically aligned channels, suppressing style-induced spurious similarity and enlarging inter-class similarity margins. To address the latter, \textbf{SHARe} reformulates localization as a hierarchical autoregressive process, injecting hierarchical ViT features in reverse depth order across refinement stages to progressively recover spatial cues lost in similarity embeddings. Experiments on COCO and Pascal VOC demonstrate substantial gains over existing methods, establishing a strong and reproducible baseline for prototype-based similarity learning in FSOD.

\newpage

\clearpage  

\section*{Acknowledgements} 
This work was supported by the National Research Foundation of Korea (NRF) grant funded by the Korea government(MSIT)(RS-2024-00456589).

%
%
\bibliographystyle{splncs04}
\bibliography{main}

\newpage

\renewcommand{\theequation}{S\arabic{equation}}
\renewcommand{\thefigure}{S\arabic{figure}}
\renewcommand{\thetable}{S\arabic{table}}

\setcounter{equation}{0}
\setcounter{figure}{0}
\setcounter{table}{0}
\setcounter{section}{0}

\renewcommand{\thesection}{\Alph{section}}
\renewcommand{\thesubsection}{\thesection.\arabic{subsection}}

\colorlet{bapbg}{Brown!15} 
\colorlet{txtplus}{Sepia!90!black} 


\title{Rethinking Prototype-based Similarity \\ Learning for Few-Shot Object Detection \\ [0.5em]
{\normalfont\Large Supplementary Material}\vspace{-0.5em}
}
\titlerunning{ReSet}
\author{}
\authorrunning{K.~Heo \textit{et al}.}
\institute{}
\maketitle


\section{Implementation Details}

\subsection{Text-to-Vision Mapping in TSMa}
We note that the mapping function $\psi$ used in \textbf{TSMa} is trained exclusively on the \textit{training split} of each dataset, which comprises only \textit{base classes} with no exposure to \textit{novel classes} — for both COCO~\cite{lin2014microsoft} and Pascal VOC~\cite{everingham2010pascal}. This ensures that $\psi$ generalizes to novel classes without any novel-class supervision when deployed in TSMa.

We adopt the Talk2DINO~\cite{barsellotti2025talking} framework, which learns a lightweight nonlinear projection $\psi: \mathbb{R}^{D_t} \rightarrow \mathbb{R}^{D_v}$ mapping CLIP~\cite{radford2021learning} text embeddings into the DINOv2~\cite{oquab2023dinov2} patch embedding space. While CLIP provides strong global image-text alignment, it lacks patch-level spatial grounding; DINOv2, on the other hand, exhibits inherent localization capability through its self-supervised patch embeddings. Talk2DINO bridges these two spaces by training only $\psi$ while keeping both encoders frozen, parameterizing the projection as a nonlinear warping of two affine transformations with a $\tanh$ nonlinearity:
\begin{equation}
    \psi(\mathbf{t}) = \mathbf{W}_b^\top \left( \tanh\left( \mathbf{W}_a^\top \mathbf{t} + \mathbf{b}_a \right) \right) + \mathbf{b}_b,
\end{equation}
which enables more flexible alignment between the two feature spaces compared to a simple linear projection.

To obtain a meaningful alignment target during training, Talk2DINO extracts per-head self-attention maps $\mathbf{A}_i \in \mathbb{R}^{H_P \times W_P}$ ($i = 1, \ldots, N$) from the last layer of DINOv2, where each head emphasizes semantically distinct regions. A head-specific region summary embedding is then computed as an attention-weighted average of the dense feature map $\mathbf{v} \in \mathbb{R}^{H_P \times W_P \times D_v}$:
\begin{equation}
    \mathbf{v}^{\mathbf{A}_i} = \sum_{h,w} \mathbf{v}[h,w] \cdot \mathrm{softmax}(\mathbf{A}_i)[h,w].
\end{equation}
Rather than averaging across heads, the head most aligned with the text embedding is selected via cosine similarity:
\begin{equation}
    \mathrm{sim}(\mathbf{v}^{\mathbf{A}_i}, \mathbf{t}) = \frac{\mathbf{v}^{\mathbf{A}_i} \cdot \psi(\mathbf{t})}{\|\mathbf{v}^{\mathbf{A}_i}\| \, \|\psi(\mathbf{t})\|},
\end{equation}
\begin{equation}
    \tilde{\mathbf{v}} = \mathbf{v}^{\mathbf{A}_j}, \quad j = \arg\max_{k=1,\ldots,N} \mathrm{sim}(\mathbf{v}^{\mathbf{A}_k}, \mathbf{t}),
\end{equation}
This max-head selection ensures that alignment is driven by the most text-relevant spatial region rather than a uniform aggregation over all heads.

Finally, $\psi$ is optimized via a symmetric InfoNCE objective over batches of image-text pairs $\{(I_i, T_i)\}_{i=1}^{B}$:
{\small
\begin{equation}
    \mathcal{L}_{\mathrm{InfoNCE}} = -\frac{1}{2B}\sum_{i=1}^{B} \log \frac{\exp(\mathrm{sim}(\tilde{\mathbf{v}}_i, \mathbf{t}_i))}{\sum_{j=1}^{B}\exp(\mathrm{sim}(\tilde{\mathbf{v}}_j, \mathbf{t}_i))} - \frac{1}{2B}\sum_{i=1}^{B} \log \frac{\exp(\mathrm{sim}(\tilde{\mathbf{v}}_i, \mathbf{t}_i))}{\sum_{j=1}^{B}\exp(\mathrm{sim}(\tilde{\mathbf{v}}_i, \mathbf{t}_j))}.
\end{equation}
}
This training requires only image-text pairs without pixel- or box-level annotations, and the resulting projected text embedding $\psi(\mathbf{t}_c)$ is directly comparable to DINOv2 patch features, enabling meaningful channel-wise interaction between text and visual features in TSMa.

\subsection{Architectural Details of SHARe}
\noindent\textbf{Stage and Layer Configuration.}
SHARe consists of $S = 5$ regression stages.
The DINOv2~\cite{oquab2023dinov2} ViT-L backbone has 24 transformer layers (indexed 0--23),
which are partitioned into five stage-aligned groups:
$\mathcal{G}_1 = \{0,\ldots,4\}$,
$\mathcal{G}_2 = \{5,\ldots,9\}$,
$\mathcal{G}_3 = \{10,\ldots,14\}$,
$\mathcal{G}_4 = \{15,\ldots,19\}$,
$\mathcal{G}_5 = \{20,\ldots,23\}$.
Note that $\mathcal{G}_5$ contains 4 layers while all other groups contain 5,
as 24 is not divisible by 5.

\noindent\textbf{Learnable Weighted Aggregation.}
Within each group $\mathcal{G}_s$, layer features are aggregated via a softmax-normalized weighted sum (Eq.~(9)). The learnable weights $w_{s,\ell}$ are initialized to zero, so each group begins as a uniform average and learns to upweight informative layers during training.

\noindent\textbf{RoIAlign.}
At each stage, RoIAlign extracts a region-specific visual cue from $\mathbf{V}_{\pi(t)}$ using a shared RoI $\mathcal{R}$, producing a spatial feature map of size $k \times k = 20 \times 20$. Rather than using the RPN proposal box directly, $\mathcal{R}$ is obtained by
expanding the proposal by at least a factor of 0.4 relative to its shorter side, ensuring sufficient context beyond the object boundary. This expansion naturally induces a binary mask $\mathbf{m}_0 \in \{0,1\}^{k \times k}$, where each cell indicates whether it falls inside the original proposal box; $\mathbf{m}_0$ serves as the initial mask input to Stage 1, and the ground-truth region mask for auxiliary supervision is derived analogously by projecting the ground-truth box onto the same expanded grid.

\noindent\textbf{Stage Condition Injection.}
In Eq.~(12), the gate $\sigma(\mathbf{m}_{t-1})$ is detached from the computation graph, preventing gradients from flowing back through the gate. This stabilizes training by decoupling the gating signal from the embedding update. The learnable injection scale $\gamma_t$ is initialized to 1 for all stages.

\noindent\textbf{Inference.}
At inference, the stage condition $\mathbf{C}_t$ is computed independently per class by expanding the RoI tensor along the class dimension, so each of the $|\mathcal{C}_{\mathrm{active}}|$ class hypotheses undergoes the full $S$-stage refinement in parallel.

\subsection{Training Losses}
The total training loss comprises three groups of terms.

\noindent\textbf{Classification.}
Per-class scoring is supervised by a focal loss applied to the logits of all active classes and one background class:
\begin{equation}
    \mathcal{L}_{\mathrm{cls}} = -\sum_{i} w_i (1 - p_i)^{\gamma} \log p_i,
\end{equation}
where $p_i$ denotes the predicted probability for class $i$, $\gamma{=}0.5$, and $w_i{=}1.5$ for the background class and $w_i{=}1.0$ for foreground classes.

\noindent\textbf{Bounding Box Regression.}
Final box coordinates are supervised with Smooth-$\ell_1$ loss between the predicted box $\hat{\bm{\delta}}$ and ground-truth box $\bm{\delta}^*$:

\begin{equation}
    \mathcal{L}_{\mathrm{box}} =
    \mathrm{SmoothL1}(\hat{\bm{\delta}}, \bm{\delta}^*).
\end{equation}

\noindent\textbf{SHARe Auxiliary Losses.}
At each regression stage $t \in \{1,\ldots,S\}$, the predicted mask logit $\mathbf{m}_t$ and the resulting region coordinate $\hat{\mathbf{r}}_t$ are supervised by four auxiliary losses. The mask is supervised by BCE and Dice losses against the ground-truth binary
region mask $\mathbf{m}^*$ derived from the expanded RoI grid:
\begin{equation}
    \mathcal{L}_{\mathrm{bce}}^{(t)} = -\frac{1}{k^2}\sum_{i}
    \left[ m^*_i \log \sigma(m_i) + (1-m^*_i)\log(1-\sigma(m_i)) \right],
\end{equation}
\begin{equation}
    \mathcal{L}_{\mathrm{dice}}^{(t)} =
    1 - \frac{2\sum_i \sigma(m_i)\, m^*_i + 1}
             {\sum_i \sigma(m_i) + \sum_i m^*_i + 1}.
\end{equation}
The predicted region coordinate $\hat{\mathbf{r}}_t \in \mathbb{R}^4$ (in $cx, cy, w, h$ format), obtained via the spatial integral layer~\cite{zhang2023detect}, is further supervised by $\ell_1$ and GIoU losses against the ground-truth region coordinate $\mathbf{r}^*$:
\begin{equation}
    \mathcal{L}_{\ell_1}^{(t)} = \|\hat{\mathbf{r}}_t - \mathbf{r}^*\|_1,
    \qquad
    \mathcal{L}_{\mathrm{GIoU}}^{(t)} = 1 - \mathrm{GIoU}(\hat{\mathbf{r}}_t,\, \mathbf{r}^*).
\end{equation}
The total loss is:
\begin{equation}
    \mathcal{L} = \mathcal{L}_{\mathrm{cls}}
    + \mathcal{L}_{\mathrm{box}}
    + \sum_{t=1}^{S}\left(
        \mathcal{L}_{\mathrm{bce}}^{(t)}
        + \mathcal{L}_{\mathrm{dice}}^{(t)}
        + \mathcal{L}_{\mathrm{\ell_1}}^{(t)}
        + \mathcal{L}_{\mathrm{GIoU}}^{(t)}
    \right).
\end{equation}
All auxiliary loss terms are weighted equally with no additional scaling coefficients.

\subsection{Entire Localization Procedure}
As described in Sec.~3.3 of the manuscript, SHARe performs localization as a hierarchical autoregressive process that progressively refines bounding boxes across $S$ stages via reverse-ordered ViT feature injection. Algorithm~\ref{alg:share} presents the complete end-to-end localization procedure, covering similarity embedding construction, RoI expansion,  stage-aligned autoregressive refinement, and final box prediction, with explicit distinction between training and inference behaviors.

\begin{algorithm}[H]
\caption{Entire Localization Procedure}
\label{alg:share}
\begin{algorithmic}[1]
\Require Query patch tokens $\{\mathbf{X}^\ell\}_{\ell=0}^{L-1}$,
         semantic prototype $\tilde{\mathbf{p}}_c$,
         RPN proposal $\mathcal{R}$,
         stage-aligned visual cues $\{\mathbf{V}_s\}_{s=1}^{S}$
\Ensure  Predicted box $\hat{\mathbf{b}}$

\Statex \textbf{// Similarity Embedding}
\State $\mathbf{f} \gets \mathrm{RoIAlign}(\mathbf{X}^{L-1},\, \mathcal{R})$
       \Comment{$k{\times}k{=}20{\times}20$ feature map}
\State $e \gets \mathrm{Linear}\!\left(\frac{\mathbf{f}}{|\mathbf{f}|_2} \cdot \frac{\tilde{\mathbf{p}}_c}{|\tilde{\mathbf{p}}_c|_2}\right)$
       \Comment{cosine similarity $\to$ embedding}

\Statex \textbf{// RoI Expansion and Initial Mask}
\State $\mathcal{R}' \gets \mathrm{Expand}(\mathcal{R},\;\texttt{min\_expansion}{=}0.4)$
\State $\mathbf{m}_0 \gets \mathbf{1}[\text{grid cell} \in \mathcal{R}] \in \{0,1\}^{k\times k}$
       \Comment{initial binary mask}

\Statex \textbf{// Stage-Aligned Autoregressive Refinement}
\State $\mathbf{e}_0 \gets e$,\quad $\mathcal{M} \gets [\mathbf{m}_0]$
\For{$t = 1$ \textbf{to} $S$}
    \State $s \gets S - t + 1$
          \Comment{reverse injection order}
    \State $\mathbf{C}_t \gets \mathrm{Linear}_t\!\left(\mathrm{RoIAlign}(\mathbf{V}_s,\, \mathcal{R}')\right)$
    \State $\mathbf{g} \gets \mathrm{StopGrad}(\sigma(\mathbf{m}_{t-1}))$
          \Comment{detached gate}
    \State $\tilde{\mathbf{e}}_{t-1} \gets \mathbf{e}_{t-1} + \gamma_t \cdot (\mathbf{C}_t \odot \mathbf{g})$
    \State $\hat{\mathbf{e}}_{t-1} \gets \mathrm{Concat}(\tilde{\mathbf{e}}_{t-1},\; \mathcal{M})$
    \State $\mathbf{e}_t \gets \mathrm{RefineBlock}_t(\hat{\mathbf{e}}_{t-1})$
    \State $\mathbf{m}_t \gets \mathrm{Conv}_t(\mathbf{e}_t)$
          \Comment{mask logit}
    \State $\mathcal{M}.\mathrm{prepend}(\sigma(\mathbf{m}_t))$

    \If{\textbf{training}}
        \State $\hat{\mathbf{r}}_t \gets \mathrm{SpatialIntegral}(\mathbf{m}_t)$
        \State Compute $\mathcal{L}_{\mathrm{bce}}^{(t)},\,
                        \mathcal{L}_{\mathrm{dice}}^{(t)},\,
                        \mathcal{L}_{\mathrm{\ell_1}}^{(t)},\,
                        \mathcal{L}_{\mathrm{GIoU}}^{(t)}$
    \EndIf
\EndFor

\Statex \textbf{// Box Prediction}
\State $\hat{\mathbf{r}}_S \gets \mathrm{SpatialIntegral}(\mathbf{m}_S)$
\State $\hat{\mathbf{b}} \gets \mathrm{RegionToBox}(\hat{\mathbf{r}}_S,\, \mathcal{R}')$

\If{\textbf{training}}
    \State \Return $\mathcal{L}_{\mathrm{box}} + \sum_{t=1}^{S}
           \left(\mathcal{L}_{\mathrm{bce}}^{(t)} + \mathcal{L}_{\mathrm{dice}}^{(t)}
           + \mathcal{L}_{\mathrm{\ell_1}}^{(t)} + \mathcal{L}_{\mathrm{GIoU}}^{(t)}\right)$
\Else
    \State \Return $\hat{\mathbf{b}}$
\EndIf
\end{algorithmic}
\end{algorithm} 
\section{Experiments}


\subsection{Computational Complexity}
Tab.~\ref{tab:tables1} reports the inference time (seconds per image) of DE-ViT~\cite{zhang2023detect}, PiDiViT~\cite{zhou2025pixel}, and our method, measured on a single RTX 3090 GPU on the COCO~\cite{lin2014microsoft} few-shot detection benchmark. Compared to the previous SOTA, PiDiViT, our method achieves a gain of \textbf{+6.6 nAP50} while running \textbf{1.6$\times$ faster} (0.71s $\rightarrow$ 0.44s). Furthermore, while PiDiViT improves over DE-ViT by only +4.0 nAP50 at the cost of 3.7$\times$ slower inference (0.19s $\rightarrow$ 0.71s), our method achieves a substantially larger gain of +10.6 nAP50 over DE-ViT with only 2.3$\times$ slower inference (0.19s $\rightarrow$ 0.44s). These results demonstrate that our method maintains efficient inference while achieving superior detection performance.

\begin{table}[t]
\centering
\caption{Inference time comparison on the COCO 30-shot setting.}
\label{tab:tables1}
\renewcommand{\arraystretch}{1.12}
\setlength{\tabcolsep}{7pt}
\resizebox{0.6\textwidth}{!}{%
\begin{tabular}{lccc}
\toprule
Method & Backbone & nAP50 ($\uparrow$) & Secs/Img ($\downarrow$) \\
\midrule
DE-ViT~\cite{zhang2023detect} & ViT-L/14 & 53.0 & \textbf{0.19} \\
PiDiViT~\cite{zhou2025pixel}  & ViT-L/14 & 58.5 & 0.71 \\
\rowcolor{bapbg}
Ours                          & ViT-L/14 & \textbf{65.1} & 0.44 \\
\bottomrule
\end{tabular}%
}
\end{table}


\begin{figure}[h]
    \centering
    \includegraphics[width=1\linewidth]{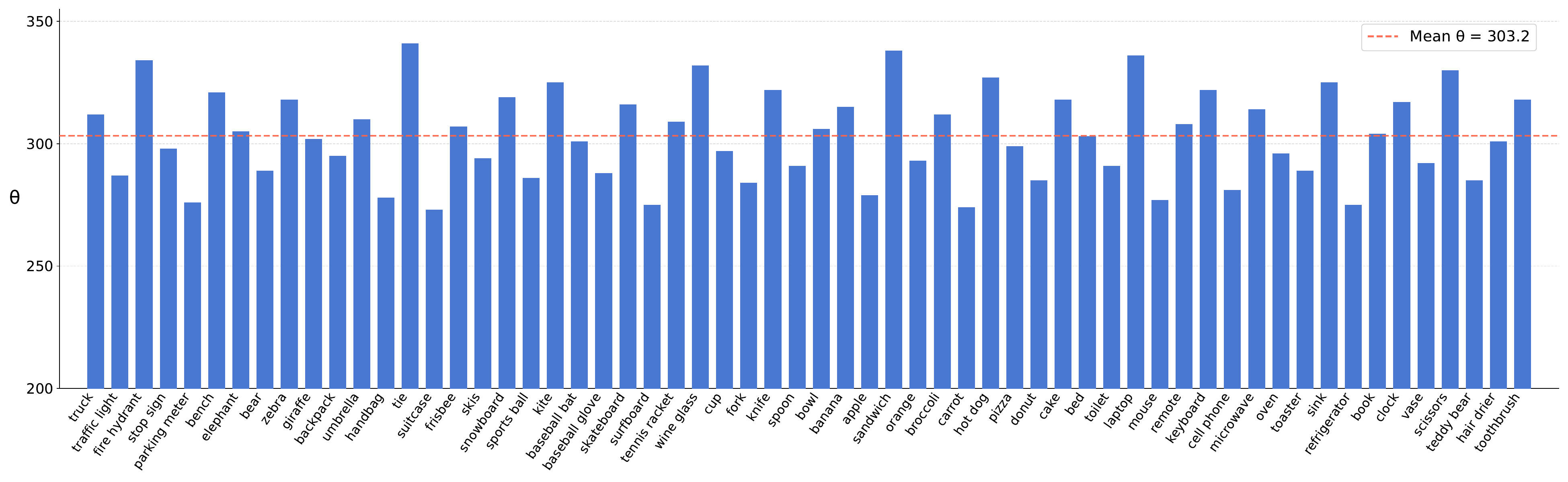}
    \caption{Distribution of learned threshold $\theta$ across base classes.}
    \label{fig:figures1}
\end{figure}

\subsection{Thresholding for Novel Classes in TSMa}
In TSMa, the channel selection threshold $\theta_c$ is learned per class during training. For \textit{base} classes, $\theta_c$ is directly optimized with sufficient labeled samples, whereas for \textit{novel} classes, direct optimization is infeasible as our framework performs no fine-tuning. We therefore set the novel-class threshold as the mean of the learned base-class thresholds. This is motivated by the empirical observation that converged $\theta_c$ values across base classes are concentrated in a narrow range (approximately 270–340, with most values near 300), as illustrated in Fig.~\ref{fig:figures1}, making the base-class mean a reasonable proxy. While per-class optimal thresholds are not available for novel categories, suppressing style-driven components and retaining class-discriminative channels remains beneficial even under this approximation, as corroborated by the analyses in Sec. 4.

\subsection{Effectiveness of TSMa under Low-Shot Settings}
While 10/30-shot settings provide a relatively sufficient number of support samples, we further analyze the effectiveness of TSMa under the more challenging 1-shot setting. Tab.~\ref{tab:tables2} reports ablation results of TSMa with performance gain $\Delta$ under both 30-shot and 1-shot (split-1/2/3/4) settings. In the 30-shot setting, adding TSMa yields a nAP50 gain of +7.9\% (60.8 $\rightarrow$ 65.6). Remarkably, in the 1-shot setting, TSMa consistently improves performance across all four splits, with gains of +8.4\%, +9.8\%, +11.5\%, and +16.6\%, respectively. These results indicate that TSMa remains highly effective even when the semantic mask is constructed from a single support image, yielding gains comparable to those observed in the 30-shot setting and demonstrating its robustness under low-shot conditions.

\begin{table}[h]
\centering
\caption{Ablation on TSMa under 1-shot and 30-shot settings.}
\label{tab:tables2}
\renewcommand{\arraystretch}{1.15}
\setlength{\tabcolsep}{7pt}
\resizebox{0.7\textwidth}{!}{%
\begin{tabular}{c ccccc}
\toprule
\multirow{2}{*}{TSMa} & \multicolumn{4}{c}{1-shot} & \multirow{2}{*}{30-shot} \\
\cmidrule(lr){2-5}
& split1 & split2 & split3 & split4 & \\
\midrule
\textcolor{black}{\xmark} & 43.0 & 43.7 & 38.1 & 32.5 & 60.8 \\
\rowcolor{bapbg}
\textcolor{black}{\cmark} & \textbf{46.6} & \textbf{48.0} & \textbf{42.5} & \textbf{37.9} & \textbf{65.6} \\
\rowcolor{bapbg}
\textcolor{txtplus}{\scriptsize \textbf{$\Delta$}} &
\textcolor{txtplus}{\scriptsize \textbf{(+8.4\%)}} &
\textcolor{txtplus}{\scriptsize \textbf{(+9.8\%)}} &
\textcolor{txtplus}{\scriptsize \textbf{(+11.5\%)}} &
\textcolor{txtplus}{\scriptsize \textbf{(+16.6\%)}} &
\textcolor{txtplus}{\scriptsize \textbf{(+7.9\%)}} \\
\bottomrule
\end{tabular}%
}
\end{table}

\subsection{K-Means Clustering in TSMa}
In TSMa, the element-wise product between visual and text feature channels is partitioned via $k$-means clustering, where we set $k=2$. This reflects the straightforward intent of separating channels into two groups: those with high and low co-activation. Using $k \geq 3$ and counting $\mathrm{count}_c(i)$ only from the highest-mean cluster would reduce the number of channels satisfying $g_{c,n}(i)=1$, excessively increasing the number of masked dimensions and consequently degrading performance.

\subsection{Aggregation Strategy in SHARe} 

\begin{table}[h]
\centering
\caption{Layer aggregation strategies in SHARe.}
\label{tab:tables3}
\renewcommand{\arraystretch}{1.12}
\setlength{\tabcolsep}{6pt}
\resizebox{0.85\textwidth}{!}{%
\scriptsize
\begin{tabular}{lcccc}
\toprule
\makecell[l]{Aggregation Strategy} & bAP & nAP & nAP50 & nAP75 \\
\midrule
Uniform Average               & 41.1 & 46.6 & 64.9 & 50.7 \\
Max Pooling                   & 41.0 & 46.2 & 65.0 & 50.5 \\
Linear Aggregation            & 40.8 & 47.0 & 65.3 & 51.0 \\
\rowcolor{bapbg}
Learnable Weighted Sum (Ours) & \textbf{41.3} & \textbf{47.4} & \textbf{65.6} & \textbf{51.5} \\
\bottomrule
\end{tabular}%
}
\end{table}

\noindent We compare four strategies for aggregating ViT layer features within each stage-aligned group $\mathcal{G}_s$ in SHARe: ``\texttt{Uniform Average}'' simply averages all layer features in the group, ``\texttt{Max Pooling}'' takes the element-wise maximum across layers at each spatial and channel position, ``\texttt{Linear Aggregation}'' computes a weighted sum of layer features with unconstrained learnable scalars, and ``\texttt{Learn-\\able Weighted Sum}'' similarly learns a per-layer scalar weight but applies softmax normalization to ensure a convex combination, initialized uniformly and optimized during training. As reported in Tab.~\ref{tab:tables3}, all strategies yield competitive and comparable performance, reflecting that any form of multi-layer aggregation provides useful stage-wise visual cues. Nevertheless, our adopted ``\texttt{Learnable Weighted Sum}'' consistently achieves the best results across all metrics, demonstrating that allowing each group to adaptively reweight its constituent layers leads to more effective stage-aligned feature injection.

\begin{figure}[h]
    \centering
    \includegraphics[width=1\linewidth]{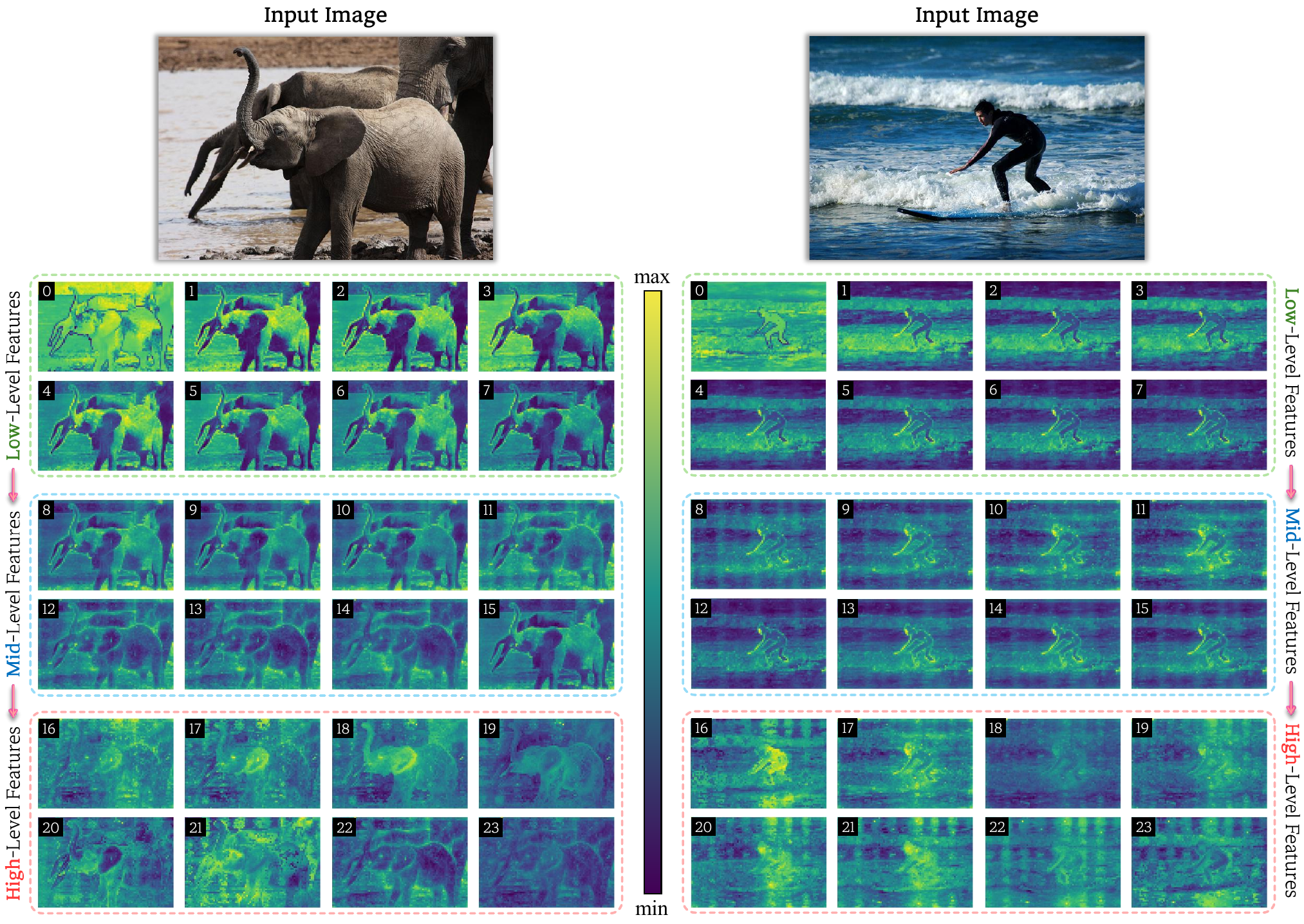}
    \caption{Layer-wise similarity maps of ViT-L features.}
    \label{fig:figures2}
\end{figure}

\subsection{Layer-wise ViT Feature Visualization}
Fig.~\ref{fig:figures2} visualizes the cosine similarity between the CLS token and patch tokens at each of the 24 layers of ViT-L, where the CLS token serves as a global representation of the image. The similarity maps reveal a transition across layers: earlier layers capture low-level cues such as edges and textures, while deeper layers increasingly encode high-level semantics and global context~\cite{dorszewski2025colors, kim2025interpreting}. This demonstrates that different layers of the ViT backbone carry distinct types of visual information, motivating the stage-aligned hierarchical feature injection in SHARe.

\subsection{Effect of the number of Stages in SHARe}

\begin{table}[h]
\centering
\caption{Effect of the number of stages $S$.}
\label{tab:tables4}
\renewcommand{\arraystretch}{1.15}
\setlength{\tabcolsep}{7pt}
\resizebox{0.55\textwidth}{!}{%
\scriptsize
\begin{tabular}{c | cccc}
\toprule
$S$ & bAP & nAP & nAP50 & nAP75 \\
\midrule
3        & 41.1 & 46.6        & \textbf{65.9} & 50.3 \\
4        & \textbf{41.4} & 46.9 & 65.8         & 51.1 \\
\rowcolor{bapbg}
5 (Ours) & 41.3 & \textbf{47.4} & 65.6       & \textbf{51.5} \\
\bottomrule
\end{tabular}%
}
\end{table}

\noindent To examine the effect of the number of regression stages $S$ in SHARe, we conduct an analysis over varying $S$ using the frozen ViT-L backbone with 24 transformer layers. For each setting, the 24 layers are evenly partitioned into $S$ stage-aligned groups of size $\lceil L/S \rceil$, each aggregated via \texttt{Learnable Weighted Sum} before stage-wise injection. As $S$ decreases, each group $\mathcal{G}_s$ spans more ViT layers, so the aggregated visual cue $\mathbf{V}_s$ mixes features from a broader range of abstraction levels rather than capturing a well-defined representation level. This blurring of the representation hierarchy weakens the stage-wise granularity of feature injection and is most clearly reflected in nAP75. As $S$ decreases from 5 to 3, nAP75 drops from 51.5 to 50.3, indicating degraded localization precision at high IoU thresholds where fine-grained spatial accuracy is critical. We adopt $S{=}5$ as it yields the best nAP and nAP75, which we attribute to its finer-grained stage alignment that best preserves the hierarchical structure of ViT representations, enabling a progressive transition from high-level semantic context to low-level spatial cues.

\subsection{Additional Quantitative Results}

\begin{table}[h]
\centering
\caption{Quantitative results across different ViT backbone sizes.}
\label{tab:tables5}
\renewcommand{\arraystretch}{1.15}
\setlength{\tabcolsep}{6pt}
\resizebox{0.95\textwidth}{!}{%
\scriptsize
\begin{tabular}{l | cccccccc}
\toprule
\multirow{2}{*}{Backbone}
& \multicolumn{4}{c}{10-shot}
& \multicolumn{4}{c}{30-shot} \\
\cmidrule(lr){2-5} \cmidrule(lr){6-9}
& bAP & nAP & nAP50 & nAP75
& bAP & nAP & nAP50 & nAP75 \\
\midrule
ViT-S/14 & 31.7 & 35.4 & 50.5 & 38.0 & 32.5 & 35.5 & 50.6 & 38.8 \\
ViT-B/14 & 38.0 & 43.4 & 62.0 & 46.6 & 38.4 & 43.5 & 62.1 & 46.7 \\
\rowcolor{bapbg}
ViT-L/14 & \textbf{41.2} & \textbf{47.1} & \textbf{65.5} & \textbf{51.5}
         & \textbf{41.3} & \textbf{47.4} & \textbf{65.6} & \textbf{51.5} \\
\bottomrule
\end{tabular}%
}
\end{table}

\noindent We report the performance of our method across different ViT backbone sizes (Small/Base/Large) under 10/30-shot settings in Tab.~\ref{tab:tables5}.

\clearpage
\subsection{Additional Qualitative Results}

\begin{figure}[h]
    \centering
    \includegraphics[width=1\linewidth]{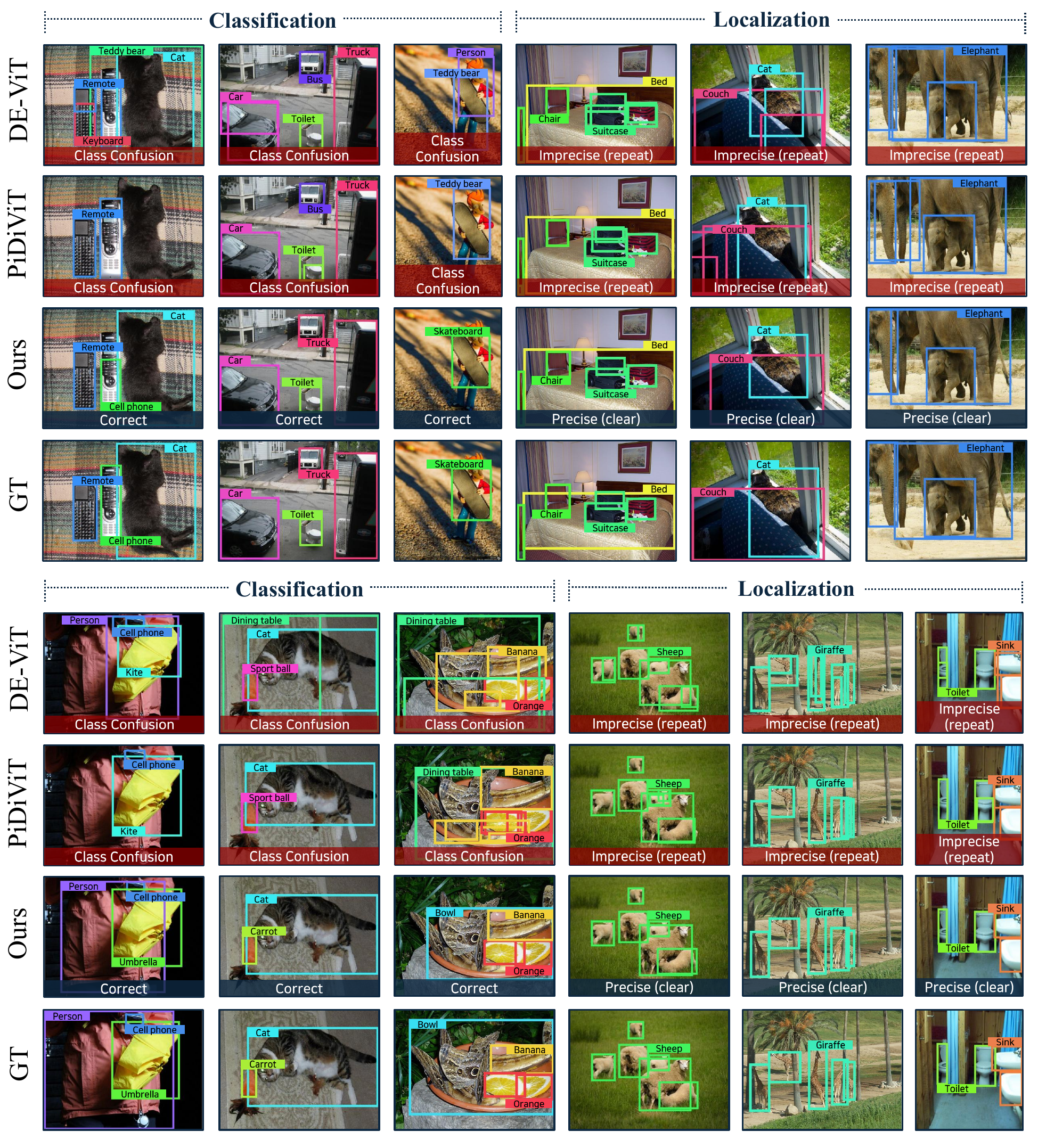}
    \caption{Qualitative comparison with existing methods (30-shot).}
    \label{fig:figures3}
\end{figure}

\end{document}